%% file: main.tex
\documentclass[10pt,twocolumn,letterpaper]{article}

\usepackage{3dv}
\usepackage{times}
\usepackage{epsfig}
\usepackage{graphicx}
\usepackage{amsmath}
\usepackage{amssymb}
\usepackage{comment}
\usepackage{multirow}
\usepackage{adjustbox}
\usepackage{rotating}

\usepackage{pifont}
\usepackage{mathtools}
\usepackage{caption}
\usepackage{subcaption}
\usepackage{float}
\usepackage[section]{placeins}

\usepackage[pagebackref=true,breaklinks=true,colorlinks,bookmarks=false]{hyperref}

\newcommand{\xmark}{\ding{55}}

\threedvfinalcopy 

\def\threedvPaperID{229} 

\begin{document}

\title{Keypoint Cascade Voting for Point Cloud Based  6DoF Pose Estimation}

\author{Yangzheng Wu  \and Alireza Javaheri \and Mohsen Zand \and Michael Greenspan \\ RCVLab, Dept. of Electrical and Computer Engineering, \\ Queen's University, Kingston, Ontario, Canada \\{\tt\small \{ y.wu, javaheri.alireza, m.zand, greenspan.michael\}@queensu.ca}}

\input{figs/fig-results}

\begin{abstract}
\vspace{-0.5\baselineskip}
We propose a novel keypoint voting 6DoF object 
pose estimation method, 
which takes
pure unordered point cloud
geometry as input without RGB information.
The proposed cascaded keypoint
voting method,
called RCVPose3D,
is based upon a novel architecture
which separates the task of semantic segmentation
from that of keypoint regression,
thereby increasing the effectiveness
of both and improving the ultimate performance.
The method also introduces a
pairwise constraint
in between different keypoints
to the loss function when regressing the quantity for keypoint estimation,
which is shown to be effective,
as well as a novel 
Voter Confident Score 
which enhances both 
the learning and inference stages.
Our proposed RCVPose3D  achieves state-of-the-art performance
on the Occlusion LINEMOD ($74.5\%$) and YCB-Video ($96.9\%$) datasets,
outperforming 
existing pure RGB and RGB-D
based methods,
as well as being competitive with RGB plus point cloud
methods.
\end{abstract}

\section{Introduction}

Accurate, robust and efficient six degree-of-freedom pose estimation 
(\emph{6DoF PE}) is an enabling technology for applications such as augmented/virtual reality~\cite{hinterstoisser2012model,posecnn}, robotic grasping~\cite{kleeberger2019large},  autonomous driving~\cite{hu2022point}, and so forth.
6DoF PE aims to
determine the rigid transformation
(comprising the 3DoF translation and 3DoF rotation)
of an object of known geometry and/or appearance within a captured scene.
This problem has been intensively investigated by the research community, initially using classical analytical approaches~\cite{aldoma2011cad,rusu2010fast,li2012worldwide}, 
and more recently exploiting the advent of machine learning 
(\emph{ML}) methods~\cite{hinterstoisser2012model,posecnn,wu2021vote,gao20206d}. 
    


    
        

A number of recent leading ML approaches~\cite{pvnet,pvn3d, wu2021vote} have been proposed based on keypoint voting, in which the 3D scene
coordinates of specific keypoints defined within an object's reference frame, are voted on and accumulated
independently for each image pixel.
The accuracy with which 
Convolutional Neural Networks (CNNs)
are able to regress geometric information
about the locations of keypoints within a scene is a main reason for the effectiveness of these approaches,
and a number of 
variations have emerged which take
both pure RGB~\cite{park2019pix2pose, yolo6d, trabelsi2021ppn, yang2021dsc, zakharov2019dpod, kehl2017ssd, posecnn, wang2021gdr}, as well as 
RGB-D~\cite{wu2021vote, shi2021stablepose, hagelskjaer2020pointvotenet, pvn3d} data
as input.

Whereas many classical approaches process 6DoF PE in 3D point clouds~\cite{rusu2010fast,TAATI2011681,li2012worldwide},
the unordered nature of point clouds presents a challenge to ML approaches for which the ordering of the data matters. 
In particular, consecutive points in the points clouds may not be in the same neighborhood and due to the lack of proximity information, they cannot be fed to CNNs 
or in general Artificial Neural Network (ANN), unlike a pixel-ordered RGB or RGB-D image.
Two approaches that have been
proposed that impose order on the
3D data are voxel-based~\cite{hagelskjaer2020pointvotenet,zhou2018voxelnet,liu2019point}, in which the point cloud is cast to a 3D grid structure, and view-based, which resamples the data from a set of directions along the view sphere~\cite{hinterstoisser2012model,zakharov2019dpod,hodan2020epos}.
A further alternative has been to consider ML
architectures that process point clouds
independent of their ordering~\cite{bruna2013spectral,fang20153d,guo20153d,masci2015geodesic,li2016fpnn,qi2016volumetric,su2015multi,wang2015voting}.
The PointNet series~\cite{qi2017pointnet,qi2017pointnet++} has emerged as the most successful such approach,
which makes use of a number of trainable
symmetric functions which are invariant to point ordering.

In this work, we 
apply the powerful keypoint voting approach, taking pure 3D 
point cloud data as input.
One challenge is
a limitation to the size of the input
point cloud. Whereas in processing conventional image-structured  data (RGB, or RGB-D), the nature of CNNs can accommodate large input images,
PointNet and other non-CNN based point cloud methods have a much stricter limit to the size of the input image.
Further, in many realistic scenes,
the object of interest comprises a 
relatively small percentage (e.g. $5\% \mbox{to} 15 \%$) of the image~\cite{hinterstoisser2012model,posecnn},
so that the majority of the input data is considered background or clutter.

To address this, 
we have developed a novel keypoint voting
architecture based on RCVPose~\cite{wu2021vote} called RCVPose3D, which partitions the
segmentation and regression tasks.
In previous work, segmentation and regression were trained and executed in parallel, whereas in the proposed architecture, which we call \emph{cascaded} keypoint voting, 
these tasks are trained 
independently and separately, and are 
executed end-to-end at inference.
This novel architecture has two main benefits: First,
training segmentation and regression separately increases the accuracy of each of these tasks,
as they each have their own independent dedicated network.
The second benefit is at inference, where background points are initially filtered out by the segmentation task,
so that only foreground points are
passed to regression.
This not only reduces the computational expense of regression, but also increases voting effectiveness, 
as background points are filtered out. Some sample results are shown in Fig.~\ref{fig:results}.

The main contributions of this work are as follows:
\begin{itemize}
\item 
A novel cascade architecture for keypoint voting based 6DoF PE that partitions the segmentation and regression tasks. This improves training of these two independent tasks, and ultimately increases performance.
Based on this architecture, we introduce RCVPose3D, which is the first variation of keypoint voting based 6DoF PE that takes pure 3D point cloud data as input.

    

\item A novel loss function that considers the Euclidean distance between pairs of simultaneously regressed keypoints as a geometric constraint, and improves keypoint estimation accuracy.

\item A novel evaluation score based on the voting space resolution. This score reduces the computational expense of training by evaluating the voter regression network, and culling certain points
before voting, 
thereby accelerating hyperparameter tuning.
    

\end{itemize}

RCVPose3D has been thoroughly evaluated and compared against other state-of-the-art (SOTA) methods, and a series of ablation studies have been performed to characterize its effectiveness.
The code is available to the public at 
\url{https://github.com/aaronWool/rcvpose3d}.

\section{Related Work}
This section reviews previous  6DoF PE work.  Tab.~\ref{tab:MethodsWithDiffInput} lists an overview of these works, classified by their input data modes of RGB, RGB-D, or 3D (i.e., point cloud).

\subsection{6DoF PE from  RGB and RGB-D Images}

Advances in 6DoF PE have been  
facilitated by the establishment of
datasets such as LINEMOD~\cite{hinterstoisser2012model} and 
YCBVideo~\cite{posecnn},
which include pose information of a variety of objects for a large number of RGB-D scenes under
various cluttered and occluded conditions.
The majority of ML-based methods have used pure RGB data images ~\cite{yolo6d,kehl2017ssd,park2019pix2pose,zakharov2019dpod,trabelsi2021ppn,yang2021dsc,wang2021gdr},
while some have
also used the depth 
(i.e. \emph{D}) information
provided by range sensors such as Microsoft Kinect~\cite{posecnn,hagelskjaer2020pointvotenet,shi2021stablepose}.
For all RGB and RGB-D methods,
the input images are multi-channel (i.e., 3- or 4-channel) 2D arrays with ordered pixel grids. These images
are naturally handled by CNNs.
OP-Net~\cite{kleeberger2020single} is notable in that it processes only the D field of an
RGB-D image, 
making use of the image
ordering in a YOLO-like grid.
\input{tabs/classified6DWorks-2}

\subsection{6DoF PE from 3D Point Clouds}
\label{sec:6DoF PE from 3D Point Clouds}
With the progress in the development
and availability of 3D acquisition sensors,
point clouds became popular in pose estimation~\cite{pvn3d,he2021ffb6d,gao20206d,hagelskjaer2020pointvotenet} and other computer vision applications~\cite{zhou2018voxelnet}. Using a point cloud, surface and geometric constraints and characteristics of a rigid object are better exploited and can improve pose estimation in certain situations~\cite{pvn3d}, such as 
in industrial applications where the scanned parts are radiometrically textureless~\cite{kleeberger2019large,hodan2017tless}. Active range sensors, such as LiDaR, are also beneficial in applications such as autonomous driving~\cite{zhou2018voxelnet,shi2020point}, where ambient lighting conditions can confound passive 2D sensors. 

In some methods~\cite{pvn3d,he2021ffb6d,wang2019densefusion},
the 
RGB data serves as the input to a CNN,
%
and the 3D geometry data only enhances feature embedding.
These methods have better performance compared to pure RGB methods, but they have never studied the impact of RGB and geometry separately.
Geometry data has also been recently used by
some RGB-D methods~\cite{zakharov2019dpod,kehl2017ssd,posecnn,wang2021gdr}
by 
applying pose refinement, 
such as ICP~\cite{Bes+McK92}, 
as a post processing step.

Prior to the advent
of ML in computer vision,
many research works were dedicated 
to the design of 
 effective 3D feature descriptors for 6DoF PE and many other applications.
This includes a large variety of 3D feature descriptors
such as Point Signatures~\cite{chua1997pointSign}, Spin Images~\cite{h2018salient},
and Point Pair Features (PPF)~\cite{drost2010ppf}. For a thorough summary of 
classical 3D feature descriptors, see~\cite{guo2016comprehensive}.
While the classical literature
was replete with 6DoF PE solutions using point cloud 
data~\cite{Guo++14},
there have been very few
such
ML-based works on 6DoF PE.
An exception is
BaseNet~\cite{gao20206d},
which makes of use of PointNet
for 3D feature extraction.

Following PointNet series~\cite{qi2017pointnet,qi2017pointnet++}, several more recent ML methods designed for 3D point cloud/mesh were introduced.
Point Transformer~\cite{zhao2021point} adapts the transformer~\cite{vaswani2017attention} to point clouds using vector attention.
DGCNN~\cite{wang2019dgcnn} uses topology-based graph convolutions to extract 3D features.
SubdivNet~\cite{hu2021subdivnet} investigate the properties of mesh and design a descriptor for it.
Lastly, PPFNet~\cite{deng2018ppfnet} embeds the classic PPF into a CNN to encode features.


In addition to those methods that apply ICP for post-processing,
there have been a few 6DoF PE methods that have 
made further use of 3D point cloud data.
PVN3D~\cite{pvn3d} fuses PointNet features with 2D convolution features to estimate keypoints, the object center, and  a semantic mask with a multi-task network loss all together. 
A least square fitting gives the final pose estimation.
In PointVoteNet~\cite{hagelskjaer2020pointvotenet} the geometry and RGB information is voxelized and an anchor box is used to localize the target object location. The final pose is given by an offset clustering.
While some recent works combine classic hand-crafted 3D descriptors such as PPF~\cite{deng2018ppfnet,deng2018ppffold} in the feature extraction stage.
others use a topology graph \cite{wang2019dgcnn,li2018pointcnn,wang2017octree} as a 3D feature descriptor instead of pure Euclidean local geometry.

\section{Background: Keypoint Voting Framework}
\label{sec:Background}
Classical voting methods such as Hough \cite{duda1972hough}, RANSAC \cite{fischler1981ransac}
Pose clustering \cite{Olson97efficientpose} and Geometric Hashing \cite{geometric-hashing} 
proved to be robust and highly effective. The accumulator space where all votes are aggregated independently, effectively filters out noise and background clutter, 
yielding an accurate estimation. With the advent of neural networks, voting techniques have gained more popularity. Recent works such as PVNet~\cite{pvnet}, PVN3D~\cite{pvn3d}, and RCVPose~\cite{wu2021vote} 
that exhibit leading SOTA performance, have merged voting-based methods, which are well established in the classical literature~\cite{wang2015voting,tejani2014latent},
with recent ML-based keypoint estimation approaches.

\input{figs/fig_framework1}
The general keypoint voting architecture
first proposed in PVNet
for RGB input,
and then modified
in PVN3D
and RCVPose
for RGB-D input,
is shown in Fig.~\ref{fig:framework1}.
We describe here the 
common elements and some variations of this architecture,
and in Sec.~\ref{sec:Approach}
we introduce a number of modifications to 
accept 
as input
unordered point cloud data.

The framework aims to estimate the 3D coordinates of a number of keypoints for an object in a scene. The keypoints themselves are simply a set of 3D points defined within the object-centric reference frame, and which therefore transform rigidly with the object.
There is no requirement that all (or indeed even any) keypoints be visible in a scene, the main criterion being that there are at least 3 keypoints per object, and that they are sufficiently separated so as to allow for accurate recovery of the object pose. Keypoints have been defined in a variety of ways, including the object bounding box corners~\cite{oberweger2018making,rad2017bb8,tekin2018real},
farthest point sampling~\cite{pvnet,pvn3d},
and disperse sampling~\cite{wu2021vote}.

As shown in Fig.~\ref{fig:framework1}, the framework passes input image $\cal{I}$ through an encoder-decoder network $ED_{SM}$. In PVNet, $\cal{I}$ was an RGB image, and $ED_{SM}$ was based on a ResNet-18 backbone. RCVPose used ResNet-152 for $ED_{SM}$
for the RGB mode of an RGB-D input image $\cal{I}$, with D being used exclusively within the loss function and only during training. In PVN3D, $\cal{I}$ was RGB-D, and there were two parallel encoder branches, with ResNet-34 applied to the RGB mode, and PointNet++ applied to the D mode. The output latent spaces from these two parallel branches were then fused and merged, and fed to the decoder stage.

The output of $ED_{SM}$ are two tensors $\cal{S}$ and $\cal{M}$, both of which have the same $W\!\!\times\!\!H$ spatial dimensions as input image $\cal{I}$. Segmentation mask $\cal{S}$ indicates to which (if any) object class each pixel belongs.

Tensor $\cal{M}$ contains the values of the regressed quantities for each pixel, that are aggregated to localize keypoints in the subsequent voting stage. 
In PVNet, $\cal{M}$ contains a set of 2D vectors for each pixel, that point in the direction of each keypoint. The subsequent voting module $V$ integrates the intersections of all such vectors in a 2D accumulator space, the peaks of which indicate the locations of keypoints (i.e., \emph{vector voting}). In PVN3D, $\cal{M}$ contains a set of 3D offsets, which translate the 3D coordinate of each pixel to vote for each keypoint within a 3D accumulator space (i.e., \emph{offset voting}). In RCVPose, $\cal{M}$ contains a set of 1D values that indicate the Euclidean distance between each pixel and each keypoint. A set of spheres with radii equal to the values in $\cal{M}$ for each pixel, are rendered within a 3D accumulator space, and the peaks at intersections of the sphere surfaces determine the keypoint locations (i.e. emph{radial voting}).

The output of voting module $V$
is tensor $\cal{T}$,
comprising $K$ 
estimated keypoints
for each of the $C$ objects.
While segmentation
provides a mechanism to handle
multiple objects,
in order to increase accuracy,
in practice
all of PVNet, PVN3D and RCVPose
consider only a single object class 
per network,
i.e. $C\!\!=\!\!1$. 
The $K\!\ge\!3$ estimated keypoint scene coordinates, along with their corresponding canonical object-frame coordinates, are passed into pose module $P$ to estimate the
object 6DoF pose
$\cal{P}$. For the purely 2D keypoint scene coordinates of PVNet, a RANSAC-based Perspective-n-Point (PnP) routine is used to recover the transformation, whereas for the 3D keypoint scene coordinates of PVN3D and RCVPose, a least squares fitting between two 3D point sets can be applied. Both PVN3D and RCVPose also refine the pose estimate further with a few ICP iterations, using a larger sample of 3D scene point data.

\section{Proposed Cascade Approach: RCVPose3D}
\label{sec:Approach}
In this work, 
we propose a novel 
keypoint-based 6DoF PE
method
to estimate the pose of an object from pure point cloud data.
The method regresses of
the radius voting quantity of RCVPose~\cite{wu2021vote},
which has been 
recently proposed.
The method combines three main 
aspects,
the first of which is the 
separation of the 
semantic segmentation and pose estimation
encoder-decoder networks,
which are arranged in a
cascade architecture (Sec.~\ref{sec:Cascade Training PE}).
The second is 
a novel loss function that considers the pairwise geometric constraints between 
simultaneously estimated keypoints (Sec.~\ref{sec:GeoConsLoss}).
The third aspect is a novel score function specific to voting methods, that facilitates training (Sec.~\ref{sec:VCS-a}).
In the following subsections, we describe these aspects in detail.

\subsection{Cascade Architecture}
\label{sec:Cascade Training PE}
The proposed cascade architecture, shown in Fig.~\ref{fig:framework2}, contains 
similar processing elements as the existing parallel architecture of Fig.~\ref{fig:framework1},
albeit in a different arrangement.
The main difference is that the segmentation and regression encoder-decoder network $ED_{SM}$ of the parallel architecture, has been decoupled in the cascade architecture
into two distinct networks,
$ED_S$ and $ED_M$.
In cascade, the point cloud is first segmented prior to being passed to the regression stage. As shown in the figure,
input point cloud $I$ first passes through the 
(now independent) encoder-decoder network $ED_S$,
which results in segmentation mask $S$. Only the filtered foreground points from $S$ are then subsequently passed to 
regression network $ED_M$, resulting in tensor $\cal{M}$ which is then passed to the subsequent voting module $V$.

This new architecture has the advantage
that each network
$ED_S$ and $ED_M$ will learn their
respective patterns independently,
rather than training them jointly in a multitask
fashion.
This may seem counter-intuitive,
as the essence of the popular multitask learning
approach is to benefit from the interaction
that occurs when training complementary tasks simultaneously. The key, however, is that
multitask learning is 
mainly beneficial 
when the tasks contain correlated information that 
reinforces the learning process~\cite{wu2021yolop,cao2017realtime,sun2020adashare}.
Our experiments (see Sec.~\ref{sec:E2Evs2S}) indicate that
segmentation and keypoint regression are sufficiently 
independent so that training them each independently significantly boosts the performance
of each network,
and ultimately the performance of pose estimation.

The other advantage of the cascade architecture is that it allows the regression network to be relatively lightweight (see Tab.~\ref{tab:e2evs2s}),
so that it is easier to train. 
Meanwhile, the estimation is typically more accurate compared to the regression existing in a parallel architecture.
Last but not least, since it is a relatively small scale network,
it will also not impact the time performance which is the initial reason most methods use parallel architecture initially.

One potential drawback of the cascade architecture is that it
requires a larger memory footprint, to accommodate both networks. The regression network in particular is relatively lightweight, as only a small proportion ($\sim 5\%$ to $15\%$) of scene points need to be accommodated,
so the memory footprint is
in practice 
only $\sim 20\%$ larger than that of the parallel network (see Sec.~\ref{sec:E2Evs2S}).

\subsection{3D Feature Extraction Backend}
\label{sec:3D Feature Extraction Backend}
In the proposed method
shown in Fig.~\ref{fig:framework2},
the encoder-decoder in the backend of 
both the segmentation network
$ED_S$
and the regression network $ED_M$,
are accomplished with an effective 3D feature extractor. The unordered nature of 3D point cloud data poses a challenge to feature extraction,
as the standard row column ordering that exists in 2D data and that is exploited by Convolutional Neural Networks is not present.
To identify an appropriate 3D feature extractor, we 
implemented a number of learnable alternatives mentioned in Sec.~\ref{sec:6DoF PE from 3D Point Clouds} and results on shown in Sec.~\ref{sec:Impact of Backend Descriptors}.

\subsection{Radial Pair Loss $\mathcal{L}_{P}$}
\label{sec:GeoConsLoss}
Existing methods
calculate loss during training
individually for each keypoint~\cite{pvnet,pvn3d,wu2021vote}.
When multiple keypoints are estimated
simultaneously, however,
the potential exists to include a loss
term that considers keypoint pairs.

Let $k_i$ be a keypoint,
with GT Euclidean distance $r_{mi}=||p_m-k_i||$
from object foreground point $p_m$.
This is the radial distance that is learned and inferred by the regression network $ED_M$. It is invariant to rigid transformations, and therefore remains constant for changes in object pose.

Let  $\hat{r}_{mi}\!=\!r_{mi}\!+\!\epsilon_{mi}$ 
be an estimate of this distance,
with residual 
$\epsilon_{mi}$.
A 
\emph{Residual Loss} term 
${\cal L}_{\epsilon}$
that considers
the residual values of only single keypoints can be formed as: 
\begin{equation}
{\cal L}_{\epsilon} = \frac{1}{M\!\times\!K}
\sum_{m=1}^M
\sum_{i=1}^{K}
SL_1(|\epsilon_{mi}|)
\label{eq:SL1}
\end{equation}
where $SL_1(\cdot)$
denotes the 
smooth L1 function,
and
which is summed
for all $M$ foreground points
over all $K$ keypoints.

An additional loss term 
is proposed here, which considers pairs of radial distances. Let 
$\Delta_{mij} = |r_{mi}\!-\!r_{mj}|$
be defined as the
\emph{radial pair difference}, which is the difference between the GT radial values from a point $p_m$ to keypoints $k_i$ and $k_j$.
Estimate
$\hat\Delta_{mij}$
of
$\Delta_{mij}$
is then formulated as:
\begin{multline}
\hat\Delta_{mij} 
=
|\hat{r}_{mi}\!-\!\hat{r}_{mj}|
= 
|r_{mi}\!+\!\epsilon_{mi} - (r_{mj}\!+\!\epsilon_{mj})| \\
\leq
|r_{mi}\!-\!r_{mj}|+|\epsilon_{mi}\!-\!\epsilon_{mj}|
\label{eq:RPD}
\end{multline}
This is an expression of the triangle inequality,
which has been applied previously to improve efficiency in
nearest neighbor search~\cite{GreenICP}. 
The radial pair difference
is complementary to
the 
magnitude of the residual used in the unary ${\cal L}_{\epsilon}$ loss.
If $\hat{r}_{mi}$ and $\hat{r}_{mj}$
are both 
either underestimates or overestimates
of their respective GT values
(i.e. 
$\text{sign}(\epsilon_{mi})\!=\!\text{sign}(\epsilon_{mj})$)
then 
$
|\epsilon_{mi} - \epsilon_{mj}| 
< 
\max(|\epsilon_{mi}|,|\epsilon_{mj}|)
$,
and the residual magnitude
dominates.
Alternately,
if 
$\hat{r}_{mi}$ and $\hat{r}_{mj}$
fall on opposite sides 
of their respective GT values,
then 
$\text{sign}(\epsilon_{mi})
= 
-\text{sign}(\epsilon_{mj})$),
and 
$
|\epsilon_{mi} - \epsilon_{mj}| 
\ge 
\max(|\epsilon_{mi}|,|\epsilon_{mj}|)
$. In this case, the
radial pair difference of
Eq.~\ref{eq:RPD} 
will exceed 
the residual magnitude.

We have exploited this by
encapsulating 
Eq.~\ref{eq:RPD}
into the following loss term,
which we call the
\emph{Radial Pair Loss}:
\begin{equation}
{\cal L}_{P} = \frac{2}{M\!\times\!K(K-1)}
\sum_{m=1}^M
\sum_{i=1}^{K}
\sum_{j=i+1}^{K}\!\!\!SL_1(
|\Delta_{mij}
-
\hat{\Delta}_{mij}|)
\label{eq:RPL}
\end{equation}


The $\cal{L_{P}}$ can be used during the whole training process,
but it can be delicate at the initial stage.
The exact same residuals in between different radii for different keypoints can exist in the initial random output.
It does become dominant
at later epochs, when the regression network training to be closer to fully convergence (i.e. the outputs approach groundtruth (GT) radii values) by enforcing the constraint on inter-keypoints distances. 
Our experiments
(see Sec.~\ref{sec:Impact of Geo Constraint Loss})
supports this premise and  shows the significant impact of
${\cal L_{P}}$
on both accuracy and training time.

\subsection{Voter Confidence Score}
\label{sec:VCS-a}
In the literature,
two classic metrics have emerged and are commonly used to evaluate the overall performance on the 6DoF PE datasets: 
ADD(S)~\cite{hinterstoisser2012model} for LMO, and ADD-S AUC~\cite{posecnn} and ADD(S) AUC for YCB.
We also introduce a new
\emph{Vote Confidence Score} (VCS) for the evaluation of the radii estimation, prior to the voting stage.

\textbf{ADD(S)} measures the average distance for asymmetric objects, and the minimum distance for symmetric objects,
between points of the object transformed with GT pose and the object transformed with the estimated pose. 
If the distance is within $10\%$ of the object diameter threshold, then the estimated pose is considered to be correct.

\textbf{ADD(S) AUC} is based on ADD(S). 
It creates a curve by plotting different thresholds against ADD(s) accuracy scores. 
The accuracy score of AUC is therefore given by the area underneath the curve.

\textbf{ADD-S AUC} is similar to ADD(S) AUC,
except it uses ADD-S only for all objects, with
the measurement based on minimum point distances.

Here we define \textbf{VCS} 
as a measure of each estimated voters' confidence level from the regression network.
VCS is formulated based on the voting space resolution. 
For a voting space resolution (e.g., edge length of a cubic voxel) of $\rho$,
a vote is considered to be correct if the absolute error between 
the GT and estimated value is less than or equal to $\rho$.
The confidence score is then the ratio of correct to total votes.
This score can estimate the
performance of the voting regression network, even before votes
are cast in accumulator space, 
and can accelerate 
hyperparameter
grid search.
For radii voting specifically, VCS is defined as:
\begin{equation}
    VCS = \frac{M'}{M}
\end{equation}
where $M$ is the number of votes and $M'$ is the number of correct radii votes when 
$\epsilon_{m,i}\!=\!|r_{m,i}\!-\!\hat{r}_{m,i}|\!<=\!\rho$.
The voting space will apparently be more confident, with sharper peaks,
when VCS is higher,
and keypoint location will thus be more accurate (see Sec.~\ref{sec:Impact of Geo Constraint Loss}).

\section{Experiments}
\label{sec:Experiments}
We evaluate the performance of our proposed RCVPose3D, and compare it with the best performing 6DoF PE methods on the two challenging 6DoF datasets that are commonly used in related SOTA work~\cite{pvnet, pvn3d}.
\subsection{Datasets}
\textbf{Occlusion LINEMOD}~\cite{hinterstoisser2012model} (LMO) is an extension of the LINEMOD dataset,
comprising 1213 annotated RGB-D images of  
9 classes of object, 
with GT pose and semantic labels. 
LMO is extremely challenging, not only because the objects within the scene are heavily occluded 
but also because it is purely for testing purposes, the convention being to train on the original LINEMOD dataset, 
which only comprises non-occluded objects. 

\textbf{YCB-Video} (YCB), is a video-based 6DoF pose dataset, initially proposed by 
PoseCNN~\cite{posecnn},
which was the first CNN for 6DoF PE.
YCB contains $130K$ frames extracted from 92 videos,
with RGB images,
depth maps, and GT poses and semantic masks provided for 21 classes of objects.
The challenge of YCB is that some frames are blurred and include occluded objects.
We follow previous works~\cite{pvnet,pvn3d,wu2021vote} and 
use a train/test split of $85\%/15\%$.

\subsection{Experimental Setup}
\label{sec:Implementation Details}
To generate the 3D data, we use the camera intrinsic parameters to transform the depth maps into point clouds.
Each point cloud is then downsampled to contain a total of $N\!=\!2^{15}$ points,
which is close to the limit that our GPU could accommodate
for a reasonable batch size of eight.
Each point cloud is then recentered
based on its bounding box, and normalized based on its maximum point value.

During training, the segmentation network inputs $N$ points and estimates a semantic label for each point, indicating which object it falls on (if any).
The segmentation output is 
randomly downsampled 
based on the estimated label probabilities,
to comprise $M \le 1024$ foreground points.
The input of the regression network
is the $M$ foreground points, all of which
have the same semantic label and thus fall on the surface of the object of interest. 
When training, these points are selected by applying the GT segmentation mask of the scene, and randomly downsampling the resulting foreground points to total $M$ points.

At inference, the regression network estimates $j\!=\!3$ keypoints' radii simultaneously~\cite{wu2021vote}.
The output of the regression network is size $M\!\times\!3$, 
comprising a radius from each input object point to all 3 keypoints.

The segmentation network's 
loss function
uses a standard Binary Cross Entropy (BCE) Loss $\mathcal{L}_{bce}$, 
whereas the regression network uses a combination of traditional Smooth L1 Loss on the radii residuals $\mathcal{L}_{\epsilon}$ (Eq.~\ref{eq:SL1}), and Radial Pair Loss  
$\mathcal{L}_{P}$ (Eq.~\ref{eq:RPL}). The regression loss is then:
\begin{equation}
    \mathcal{L}_r = \alpha \mathcal{L}_{\epsilon} + \beta \mathcal{L}_{P}
\end{equation}
where $\alpha$ and $\beta$ are weights that are adjusted during the training as follows: 
During the first 100 epochs,
we set $\alpha\!=\!0.8$ and $\beta\!=\!0.2$ so that the network learns to approach the actual radii values with minor assistance from the radial pair constraint.
For the remaining 150 epochs, $\alpha$ and $\beta$ are adjusted to $\alpha\!=\!0.2, \beta\!=\!0.8$ for fine-tuning when $L_{\epsilon}$ has mostly converged and 
$\mathcal{L}_{P}$ dominates.

The Adam~\cite{Adam} optimizer is used for both networks during training.
The initial learning rate is set to $1e^{-4}$ and reduces $\times 0.1$ for every 50 epochs.
We train the networks with a batch size of $8$ for segmentation and $32$ for regression on three RTX6000 GPUs.

\input{tabs/results_mean-rev2}
\subsection{Performance Evaluation}
To our knowledge, 
BaseNet~\cite{gao20206d} is the only other ML-based 6DoF PE method that inputs unordered point cloud data. 
OP-Net~\cite{kleeberger2019large} does take pure depth data as input, although they use row-column image ordering using a YOLO-style~\cite{redmon2017yolo9000} 2D grid decomposition, on a less common dataset.
In addition to BaseNet,
we also compare RCVPose3D with SOTA RGB and RGB-D methods.
Two classic metrics are used to evaluate the overall performance of RCVPose3D, 
ADD(S)~\cite{hinterstoisser2012model} for LMO, and ADD-S AUC~\cite{posecnn} and ADD(S) AUC for YCB.
Our experiments show that RCVPose3D is competitive with these other methods, 
even though it does not make use of any RGB information.

As shown in Tab.~\ref{tab:resultscombined}, on YCB, RCVPose3D performs better than all pure RGB methods. 
Its performance 
is especially relatively strong on objects such as \emph{wood block} ($+3\%$ better) and \emph{mug} ($+1.7\%$ better) which have a uniform color (see Supp. Mat.). 
RCVPose3D also outperforms all other SOTA methods with the relatively strict ADD(S) AUC metric,
both with and without ICP refinement. 
It performs second best with the ADD-S AUC metric,
approaching
RCVPose~\cite{wu2021vote} within a small margin ($-0.6\%$).

On LMO, 
RCVPose3D again outperforms all SOTA methods that use RGB data. 
Similar to YCB, it estimates a better pose on objects with poor radiometric texture (almost uniform color) such as \emph{eggbox} and \emph{can}. It outperforms RCVPose by $+3.4\%$ for \emph{can} (see Supp. Mat.).
Compared to all methods, RCVPose3D is the second-best 
with on average only $-0.6\%$ less than PointVoteNet~\cite{hagelskjaer2020pointvotenet} which inputs both point cloud and RGB data (see Tab.~\ref{tab:resultscombined}).

Overall, RCVPose3D is the best performing metric for 6DoF PE on pure point cloud geometry. It also outperforms all methods that input pure RGB data. Finally, It is competitive with all other RGB-D/RGB+3D 6DoF PE methods, ranking either best or second best for different datasets and metrics.


\section{Ablation Studies}
\label{sec:Ablation Studies}

We executed a series of experiments to characterize the
benefits of the
various novel elements of RCVPose3D
using the common framework
presented in 
Sec.~\ref{sec:Approach}.
In particular, we characterized the impact of the cascade architecture and the Radial Pair Loss.
We also compared the relative effectiveness of different backend 3D feature extractors.
These tests were all executed on the complete LMO dataset.

\subsection{Parallel vs. Cascade, and Radial Pair Loss}
\label{sec:E2Evs2S}
A set of experiments were executed to 
isolate the impact of the cascade architecture on both the segmentation and regression stages, as well as the effect of Radial Pair Loss and the overall pose estimation accuracy.

For the cascade architecture, the configuration is as  in Sec.~\ref{sec:Implementation Details}. 
The parallel architecture takes the same input of $N\!=\!2^{15}$ points.
Its output, however, is $N\!\times\!4$, which includes a semantic label and the $3$ radii values to each of the corresponding keypoints for all $N$ points. 
\input{tabs/e2evs2s_combined}
\textbf{Segmentation:} 
The first experiment investigates
segmentation in isolation.
The segmentation output 
(i.e. $\cal{S}$ of Fig.~\ref{fig:framework1})
is compared for both architectures
by measuring mIoU against GT.
As is the case with the cascade architecture, different loss functions are also applied to the parallel architectures.
Initially, the Smooth L1 Loss $\mathcal{L}_{s}$ is used to train both the segmentation and regression components,
as in PVNet~\cite{pvnet}.
In order to 
compare with the cascade architecture fairly, 
the proposed Radial Pair Loss and BCE Loss are then applied afterwards.

The results are shown 
in Tab.~\ref{tab:e2evs2scombined}.
It can be seen that  cascade significantly outperforms parallel,
by over $+15\%$ in testing. Also, the Radial Pair Loss alone boosts performance of the parallel architecture by $+13.1\%$.

\textbf{Regression:} A variation of the above experiment was
repeated to evaluate regression in isolation.
Here, the regression output
($\cal{M}$ of Fig.~\ref{fig:framework1})
is compared for both architectures
and loss function combinations,
by measuring VCS against GT.
The results in Tab.~\ref{tab:e2evs2scombined}
once again confirm that 
cascade outperforms parallel 
by $+37\%$,
and 
Radial Pair Loss boosts parallel performance
by $+9.6\%$.

It can be seen that  cascade significantly outperforms parallel,
by over $+15\%$ in testing. Radial Pair Loss alone boosts performance of the parallel architecture by $+13.1\%$.

\input{tabs/e2evs2s_occLINEMOD}
\textbf{Pose Estimation:} A further experiment evaluated the above combination of elements for the complete
6DoF PE pipeline,
the results of which are shown in 
Tab.~\ref{tab:e2evs2s}.
The overall ADD(S) is boosted
for cascade by $+13\%$ on average compared to parallel, with a small runtime sacrifice of 1 fps and a GPU memory cost of 4.7GB. 

\subsection{Comparison of Backend Feature Descriptors}
\label{sec:Impact of Backend Descriptors}
There exist various backend networks in the literature,
each encoding a distinct 3D feature descriptor.
In order to find the most effective 3D feature descriptor,
we compared the performance of RCVPose3D with different backend networks introduced in Sec.~\ref{sec:3D Feature Extraction Backend}.
The result in Tab.~\ref{tab:descriptors_combined} show that
PointNet++~\cite{qi2017pointnet++} is the most accurate feature extractor,
with Point Transformer~\cite{zhao2021point}
as a close second.
PointNet was also more efficient,
requiring only 1.8M parameters,
with the next smallest being Point Transformers
with 4.7M.
We therefore used PointNet++ in the final 
RCVPose3D configuration, 
and for all other ablation experiments.

\input{tabs/desciptors_combined}

\subsection{Impact of Number of Votes}
\label{sec:NoofVotes}
\input{figs/fig-NoofVotes}
In order to justify the 
number $M$ of foreground points
fed into the regression network,
which in turn leads to $M$
votes in the subsequent voting module, 
we trained the network with different numbers of votes forwarded through regression and into the accumulator space.
Here
we used the
LMO \emph{ape} object,
and varied the number of votes over a range of
from $M=2^{7}$
to $2^{11}$.
The remaining hyperparameters 
were fixed,
except we reduced the batch size from $8$ to $4$ when $M=2^{11}$, due to GPU memory limits.  

Fig.~\ref{fig:NoofVotes} shows the accuracy for different number of votes using different metrics.
The training VCS is consistently accurate and only drops slightly for $2^{11}$ votes.
However, the testing VCS is significantly worse when there are fewer points.
ADD(S) also shows a similar trend.
Both VCS and ADD(S) slightly decrease when the number of votes is larger than $2^{10}$. 
This is most likely due to the smaller batch size for $2^{11}$ votes.
We ended up using $M=2^{10}$ votes for optimal performance.
\subsection{Impact of Radial Pair Loss $\mathcal{L}_{P}$}
\label{sec:Impact of Geo Constraint Loss}
In order to show the impact of the proposed $\mathcal{L}_{P}$, we trained a radii regression network with the same configuration in Sec.~\ref{sec:Implementation Details} but
using $\mathcal{L}_{\epsilon}$, and compared it with the combined $\mathcal{L}_{\epsilon} + \mathcal{L}_{P}$ loss.
We then use VCS, the newly proposed score proved to be consistent with ADD(S) in Sec.~\ref{fig:NoofVotes} and Sec.~\ref{sec:E2Evs2S}, to evaluate the regression performance. 
As shown in Tab.~\ref{tab:geoloss}, the network supervised by $\mathcal{L}_{P}$ is not only more accurate, but it also converges faster.
Performance actually doubles for learning rate $1e\!-\!3$ in which the network does not perform well using 
$\mathcal{L}_{\epsilon}$.
\input{tabs/geoLoss}

\section{Conclusion}
We propose a novel cascade architecture for 6 DoF PE from pure point cloud data, i.e. without RGB information. 
PointNet++~\cite{qi2017pointnet++} is selected for the segmentation and regression backbones, after carefully considering different 3D feature extractors.
A novel radial pair loss function ($\mathcal{L}_{P}$) is proposed and shown to further improve the performance. Finally, a novel score, Vote Count Score (VCS), for the accuracy of the regression network voting technique (VCS) is proposed and shown to improve training.
We achieve competitive results on two popular datasets, LMO and YCB, where RCVPose3D outperforms the SOTA methods that use pure RGB information. When compared to methods that input RGB-D and RGB + point cloud data, RCVPose3D is the best performing method in YCB with the ADD(S) AUC metric. It is also the second best for other combinations of datasets and metrics.
Lastly, our time performance is 15 fps, 
which is competitive with other SOTA methods.

\noindent{\bf Acknowledgements:} This work was supported by Bluewrist Inc. and NSERC.

{\small
\bibliographystyle{ieee_fullname}
\bibliography{egbib}
}

\end{document}


\title{Supplementary Material:\\
Keypoint Cascade Voting for Point Cloud Based  6DoF Pose Estimation}

\author{Yangzheng Wu  \and Alireza Javaheri \and Mohsen Zand \and Michael Greenspan \\ Dept. of Electrical and Computer Engineering, 
Ingenuity Labs Research Institute \\ Queen's University, Kingston, Ontario, Canada \\{\tt\small \{ y.wu, javaheri.alireza, m.zand, greenspan.michael\}@queensu.ca}}

\maketitle
\section{Overview}
Here we show additional experiments and their associated results. 
First, the accuracy of RCVPose3D for each individual object is compared with other methods in Sec.~\ref{sec:perObject} as well as samples for all LMO objects in Fig.~\ref{fig:resultsmore}. 
In Sec.~\ref{sec:Best 3D Feature Extractor}, more experimental results are provided to support our decision on choosing PointNet++ as the backend feature extractor in RCVPose3D. 
We report detailed results in Tab.~\ref{tab:YCBVideoDescriptors} and Tab.~\ref{tab:OccLinemodDescriptors},
with time performance listed in Tab.~\ref{tab:convergenceAndTParam}.


\section{Accuracy Results per Object\label{sec:perObject}}

In Sec.~\textcolor{red}{5.3} of the main paper, we have claimed that the performance of RCVPose3D is relatively strong on objects with uniform colors, such as \emph{wood block} and \emph{mug} in YCB. The performance gains for these objects are respectively $3\%$ and $1.7\%$. In Tab.~\ref{tab:YCBVideoFull}, AUC and ADD(s) results of different methods are listed for all the objects in the YCB dataset. A similar statement was made for \emph{eggbox} and \emph{can} on LMO. Tab.~\ref{tab:OccLinemod} shows that RCVPose3D outperforms other methods by $3.4\%$ for \emph{can}, and hence, it is the second best by only $0.6\%$ lower ADD(s) for \emph{eggbox}. 

\section{Best 3D Feature Extractor\label{sec:Best 3D Feature Extractor}}
In order to identify the best 3D feature extractor for RCVPose3D, we 
implemented and/or tested a number of learnable feature extractors including PointNet++~\cite{qi2017pointnet++}, Point Transformer~\cite{zhao2021point}, DGCNN~\cite{wang2019dgcnn}, SubdivNet~\cite{hu2021subdivnet}, and L-PPF \& L-FPFH.


\textbf{1) PointNet++~\cite{qi2017pointnet++}} 
constructs a series of feature vectors with a Set Abstraction (SA) Module, which groups the points and encodes local geometry into learnable features. 
We used three Set Abstraction layers.
Farthest Point Sampling 
is used to partition the point cloud into
a set of clusters,
which covers the entire point cloud.
It samples cluster centers and 
ensures the points are sparse 
to avoid overlapping between the clusters. It
therefore results in sampling the entire point cloud properly.
We set two levels of scales in Multi-Scale Grouping (MSG) of each Set Abstraction Layer.
A series of 2D convolution layers with batch normalization and max pooling are applied on the grouped points 
feature vector 
to generate a convolutional feature vector associated with each cluster center.
The PointNet++ descriptor creates four different hierarchy clusters which encode both local and global geometries.
It is also called Vanilla since it is the most original design of Pointnet series.

\textbf{2) Point Transformer~\cite{zhao2021point}} is an extended version of Transformers performed on Natural Language Processing (NLP) and visual processing tasks. 
Leaving out the unordered nature of point clouds, point cloud structure is very similar to linguistic data,
in that the self attention mechanism still applies to point clouds when encoding features from different positions.

According to~\cite{zhao2021point}, there are two types of attention in literature, i.e. \emph{scalar} and \emph{vector}.
Vector attention is more accurate since it adapts itself to all feature channels separately and not to the whole feature vectors as in scalar attention~\cite{zhao2021point}.
We used the semantic segmentation structure of the Point Transformer to segment the point cloud in our architecture.
The regression part is also modified accordingly in order to produce pointwise radii estimation.

\textbf{3) DGCNN~\cite{wang2019dgcnn}} encodes the feature descriptor by using a k-nearest neighbor (KNN) approach.
For each point in the input point cloud of size $N\!\times\!3$, 
DGCNN finds the $k\!=\!20$ nearest neighbor points and 
concatenates indices of those points together to encode the topological graph of the local region. 
Then a 2D convolution is applied to the graph and encodes it into feature vectors. 
This process is repeated multiple times during the feature extraction stage. 
DGCNN is a topology graph-based encoder that encodes only local features. 

\textbf{4) Subdivnet~\cite{hu2021subdivnet}} is a network designed for encoding triangle meshes by simulating a 2D image convolution.
The convolution kernel operates on a neighborhood of four manifold triangles. 
The triangle 
$e$ 
at the center is the kernel of 
a set of neighboring triangles, denoted as {$e_{0}$, $e_{1}$, ..., $e_{K-1}$}.
The mesh convolution kernel is defined by a linear combination of four elements, i.e.,
the
center feature, the sum of neighbors, the differential sum in between neighbors, and the differential sum in between neighbors and the center.
It is formulated as: 
\begin{equation}
\begin{array}{rl}
    Conv(f) =& w_{0}e + w_{1}\sum\limits_{i=0}^{K-1}e_{i} + \\
&w_{2}\sum\limits_{i=0}^{K-2}|e_{i+1}-e_{i}| + \\
&w_{3}\sum\limits_{i=0}^{K-1}|e-e_{i}|
\end{array}\label{eq:subdivnet}
\end{equation}
where $w_{0},w_{1},w_{2},w_{3}$ are trainable weights. 
We use a kernel size of $3$, so that the sum is from the three neighbors of the center mesh.
The features are extracted by mesh convolutions and encoded to be similar to an 2D image.

\textbf{5) L-PPF \& L-FPFH} is
inspired by PPFNet~\cite{deng2018ppfnet} and DGCNN~\cite{wang2019dgcnn}, 
and uses a K-Nearest Neighbor (KNN) approach to encode local neighborhoods.
Two classic feature descriptors, FPFH\cite{rusu2009fpfh} and PPF\cite{drost2010ppf}, are interpolated with learnable weights to assist the encoder of the network.
The encoder contains multiple hybrid convolutional blocks.
Each block contains three layers in general.
The first layer takes pure point cloud as input.
Then, the Farthest Point Sampling
selects the cluster centers $p_{i}$.
For each $p_{i}$,
$k$ nearest neighbors are selected.
Since $k$ nearest points are forced to be selected, it is robust to 
uneven point distribution 
on the surface of the object.
We concatenate these hand-crafted descriptors to the network.
Farthest Point Sampling is used to sample the input $N \times 3$ points into $M$ clusters centered at 
$P^{c}=\{p^{c}_{i}|i=0,1\cdots M\}$ 
$
P^{c}\!=\!\{p^{c}_{i}\}_{i=1}^{M}
$
initially within each block.
Then the $K$ nearest neighbor points 
$P^{k}=\{p^{k}_{i}|i=0,1\cdots K\}$
$
P^{k}\!=\!\{p^{k}_{i}\}_{i=1}^{K}
$
within a certain range of $d$ of the cluster centers
are selected to compute the classic descriptor.
PPF is based on the point pairs in between each cluster center $p^{c}_{i}\!\in\!P^{c}$ and those points $p^{k}_{i}\!\in\!P^{k}$ among them.
Similarly, FPFH is encoded based on all points $p^{k}_{i}\!\in\!P^{k}$ within each of the clusters $p^{c}_{i}\!\in\!P^{c}$.
The $k\!\times\!f$ classical descriptor vector (in the L-PPF scenario, $f\!=\!4$) is then flattened and stacked into a 
$m\!\times\!(k\!\times\!f)$ 
classical feature vector.
To make it learnable, a 1D convolution with a kernel size of $f$ is applied to this feature vector.
The exact same operations are duplicated multiple times, and 
each time the cluster centers are sampled based on the previous layers' centers.
In this way, features at different hierarchies are encoded into the network.

These descriptors are further adjusted to fit as the feature extraction backend in the RCVPose3D architecture.
All descriptors are trained with the identical hyperparameter configurations mentioned in Sec.~\textcolor{red}{5.2}.
It was shown in Sec.~\textcolor{red}{6.2} of the main paper that PointNet++~\cite{qi2017pointnet++} and Point Transformer~\cite{zhao2021point} are the most accurate 3D feature descriptors compared to the other alternatives. Tab.~\ref{tab:OccLinemodDescriptors} and Tab.~\ref{tab:YCBVideoDescriptors} contain the same results per object.
PointNet++ is also the fastest~(w.r.t. training) and the most memory efficient feature descriptor.
It converges in $16h$ and requires only $1.8M$ parameters, 
compared to $22h$ convergence time and $4.7M$ parameters of the Point Transformers in the second place, as shown in Tab.~\ref{tab:convergenceAndTParam}.
\input{tabs/convergenceTrainableParam}

\clearpage
\input{tabs/occLINEMOD-rev2}
\input{tabs/occlinemod_descriptors-rev2}
\input{tabs/YCB}

\input{tabs/ycb_descriptors}
\input{figs/fig-resultsmore}
\clearpage
{\small
\bibliographystyle{ieee_fullname}
\bibliography{egbib}
}

%% file: figs/fig-results.tex
\twocolumn[{
\renewcommand\twocolumn[1][]{#1}%
\maketitle
\vspace{-2\baselineskip}
\begin{center}
  \includegraphics[width=\textwidth]{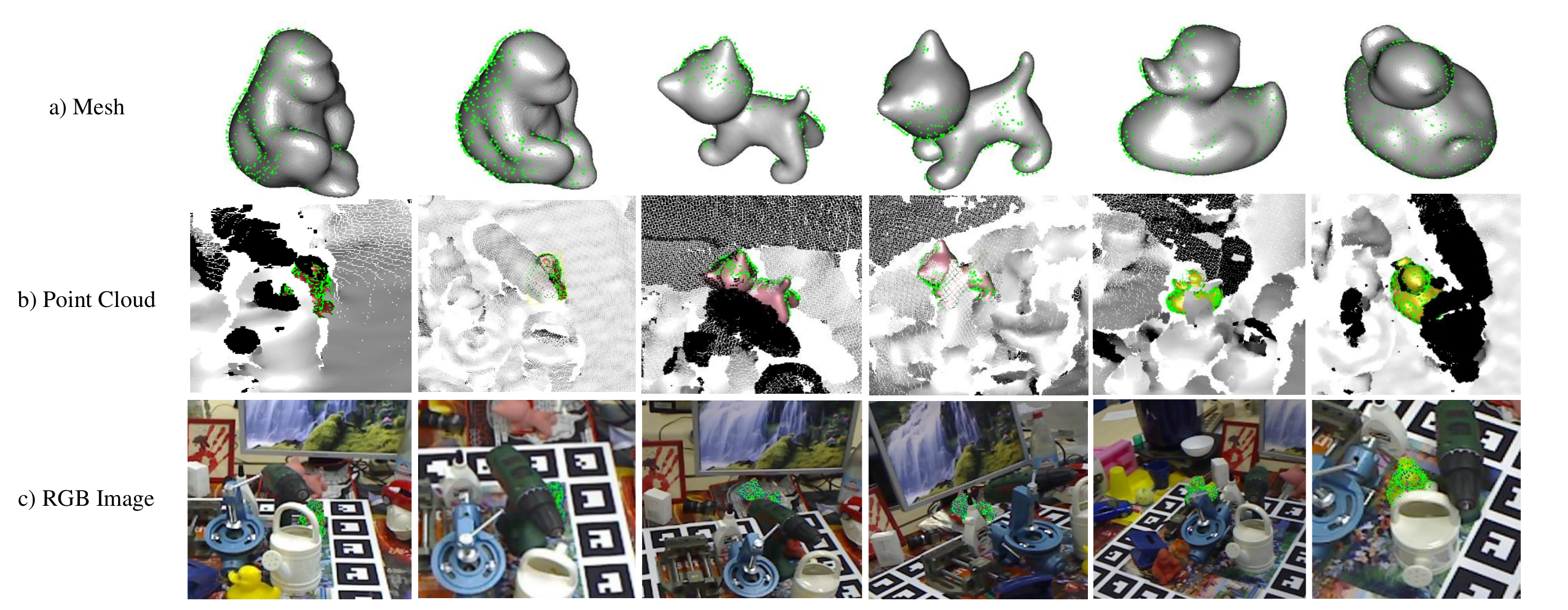}
  \vspace{-1.5\baselineskip}
  \captionof{figure}{Samples of RCVPose3D results on Occlusion LINEMOD: The meshes are applied with groundtruth (GT) pose, the green points are applied with estimated poses, whereas the blue dots are projected GT poses. The color in Point Cloud and RGB images are for illustration only, as RGB data is not used for training or inference} 
  \label{fig:results}
\end{center}
\vspace{0\baselineskip}
}]

%% file: tabs/classified6DWorks-2.tex
\begin{table}[t]
\begin{center}
\begin{tabular}{ccccc}
\cline{1-5}
\multicolumn{1}{c}{\multirow{2}{*}{Method} } &
\multicolumn{1}{c}{Publication} &\multicolumn{3}{c}{Data Mode} \\
& Date & RGB & D & 3D \\
\hline \hline
Tekin et al. \cite{yolo6d} & 2017 & \checkmark & & \\  
SSD-6D \cite{kehl2017ssd}& 2017  & \checkmark &  & \\
Pix2Pose \cite{park2019pix2pose} & 2019 & \checkmark & &  \\ 
DPOD \cite{zakharov2019dpod}& 2019 & \checkmark &  &  \\ 
Trabelsi et al. \cite{trabelsi2021ppn}& 2021 & \checkmark & &  \\ 
Dsc-posenet \cite{yang2021dsc} & 2021 & \checkmark & & \\ 
GDR-Net \cite{wang2021gdr}& 2021 & \checkmark &  & \\
\hline

PoseCNN \cite{posecnn}& 2017 & \checkmark & \checkmark & \\ 
Tian et al. \cite{tian2020robust}&2020& \checkmark & \checkmark  &\\

StablePose \cite{shi2021stablepose}& 2021 & \checkmark & \checkmark & \\  
SO-Pose \cite{Di_2021_ICCV}&2021& \checkmark & \checkmark  &\\
\hline

PVN3D \cite{pvn3d} & 2019 & \checkmark & & \checkmark \\
DenseFusion \cite{wang2019densefusion}& 2019 & \checkmark & & \checkmark \\ 
PointVoteNet \cite{hagelskjaer2020pointvotenet}& 2020 & \checkmark & & \checkmark \\ 
FFB6D \cite{he2021ffb6d}& 2021 & \checkmark & & \checkmark \\\hline

RCVPose \cite{wu2021vote}& 2021 & \checkmark & \checkmark & \checkmark \\ \hline
OP-Net~\cite{OPNet} & 2019 & & \checkmark &  \\\hline
BaseNet~\cite{gao20206d} & 2020 & & & \checkmark \\
RCVPose3D& 2022 & & & \checkmark \\ \hline
\end{tabular}
\end{center}
\vspace{-1\baselineskip}
\caption{Methods with various input data modes}
\vspace{-1\baselineskip}
\label{tab:MethodsWithDiffInput}
\end{table}

%% file: figs/fig_framework1.tex
\begin{figure}[t]
\begin{subfigure}{\columnwidth}
\begin{center}
    \includegraphics[width=0.95\textwidth]{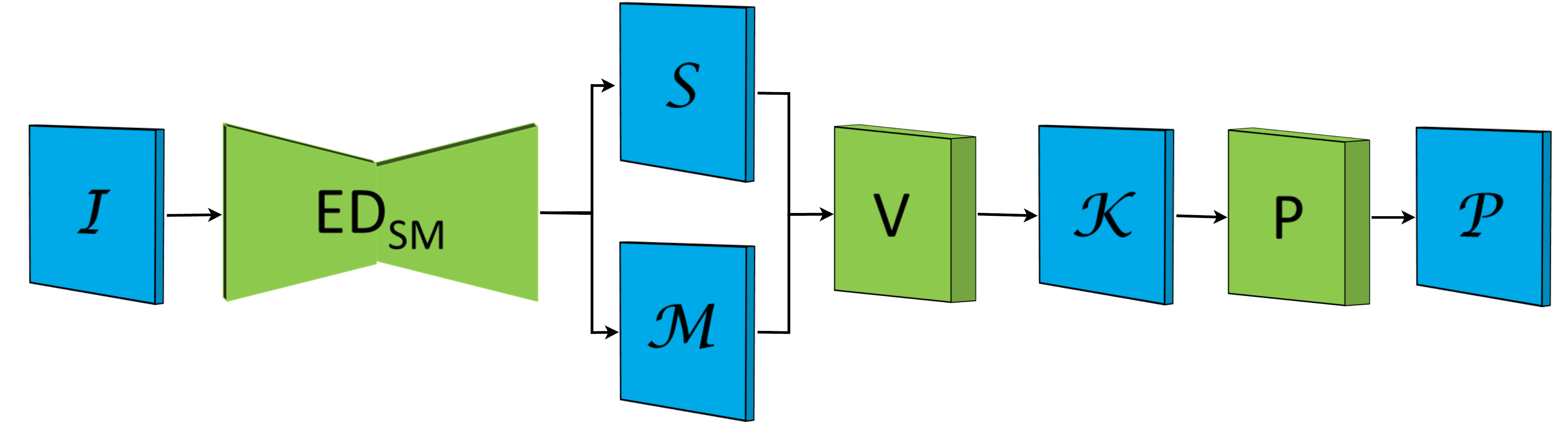}
    \caption{Existing Parallel Architecture}
  \label{fig:framework1}
  \end{center}
\end{subfigure}

\vspace{\baselineskip}

\begin{subfigure}{\columnwidth}
    \begin{center}
    \includegraphics[width=0.95\textwidth]{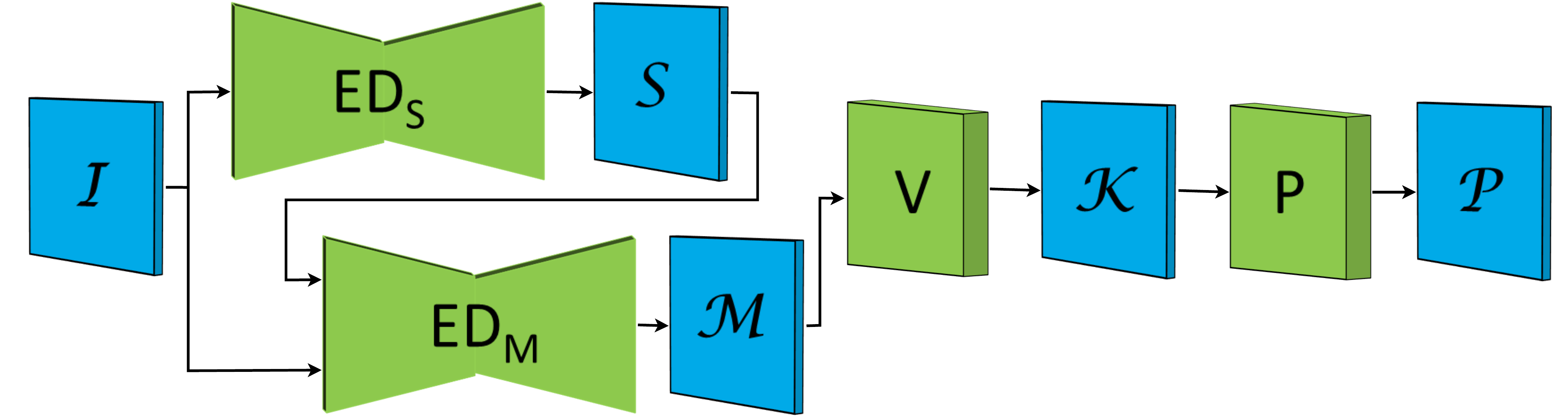}
    \caption{Proposed Cascade Architecture}
    \label{fig:framework2}
    \end{center}
\end{subfigure}

\caption{Existing parallel and proposed cascade keypoint voting architectures: $I$=input, $ED_{SM}$=combined segmentation/regression encoder-decoder,
$ED_S$=segmentation encoder-decoder,
$ED_M$=regression encoder-decoder,
$\cal{S}$=segmentation tensor, $\cal{M}$=regression tensor, $V$=voting module, $\cal{K}$=estimated keypoint coordinates, $P$=pose estimation module, $\cal{P}$=output estimated 6DoF object poses}
\vspace{-1.5\baselineskip}
\end{figure}

%% file: tabs/results_mean-rev2.tex
\begin{table}[t]
\begin{center}
\begin{adjustbox}{max width=\columnwidth}
\begin{tabular}{c|c|c|c|c|c}
\cline{4-6}
\multicolumn{3}{c|}{}

 & \multicolumn{2}{c|}{YCB} & LMO \\ \cline{3-6}
\multicolumn{2}{c|}{}&Optional

  &ADD(S) & ADD-S & $\mbox{ADD(s)}$\\ 
\cline{1-2}

Mode      & Method &Refine? &AUC &AUC& $ [\%]$
\\\hline \hline
                                & Oberweger~\cite{oberweger2018making} & \xmark  &72.8&-& 27.1\\\cline{2-6}  
                                & Hu et al.~\cite{hu2019segmentation}  &\xmark  &-&-& 30.4\\\cline{2-6}  
                                & Pix2Pose~\cite{park2019pix2pose} &  \xmark   &-&-& 32.0\\\cline{2-6}  
                                & \multirow{2}{*}{DPOD~\cite{zakharov2019dpod}}  &\xmark  &-&-& 32.8\\\cline{3-6}  
                                &  &\checkmark  &-&-& 47.3\\\cline{2-6}  
                                & PVNet~\cite{pvnet}    & \xmark   &73.4&-& 40.8\\\cline{2-6}  

\multirow{-2}{*}{RGB}                                & PPRN~\cite{trabelsi2021ppn}  & \xmark   &83.1&-& 58.4\\\cline{2-6}  
                                & SSD6D~\cite{kehl2017ssd}& \checkmark  &-&-& 27.5\\\cline{2-6}

                                &  \multirow{2}{*}{PoseCNN~\cite{posecnn}}   & \xmark  &59.9&75.8& 24.9\\\cline{3-6}
                              &  & \checkmark &85.4&93.0& - \\\cline{2-6} 

                                
                                 & GDR-Net~\cite{wang2021gdr} & \checkmark  &84.4&91.6& 62.2\\\cline{2-6} 
                                 & SO-Pose~\cite{Di_2021_ICCV} & \checkmark  &83.9&90.9& 62.3\\\hline\hline
& Tian et al.~\cite{tian2020robust}&\xmark  & - & 91.8  &-\\ \cline{2-6}  

                                & \multirow{2}{*}{RCVPose~\cite{wu2021vote}}      &  \xmark    & 95.2 & 96.6 & 70.2 \\\cline{3-6}  
\multirow{+2.5}{*}{RGB-D/}  

& &\checkmark  & \underline{95.9} & \textbf{97.2}  & 71.1\\ \cline{2-6}  
\multirow{+2.5}{*}{RGB+3D}  & \multirow{2}{*}{PVN3D~\cite{pvn3d}} &\xmark   & 91.8 & 95.5 & 63.2\\\cline{3-6}
&  &\checkmark  & 92.3  & 96.1 & -\\\cline{2-6}
& FFB6D~\cite{he2021ffb6d}&\xmark  &92.7 &\underline{96.6} & 66.2\\\cline{2-6}  
& PointVoteNet\cite{hagelskjaer2020pointvotenet}   &\checkmark &-&-& \textbf{75.1}\\\hline\hline
\multirow{2}{*}{3D/Mesh}
& \multirow{2}{*}{BaseNet~\cite{gao20206d}}  &\xmark &- &91.4 & -\\ \cline{3-6}
&  &\checkmark &- & 94.7 & -\\ \cline{2-6}
& \multirow{2}{*}{RCVPose3D}  &\xmark &95.7 &96.3 & 73.7\\ \cline{3-6}
&  &\checkmark &\textbf{96.0} &\underline{96.6} & \underline{74.5}\\ \hline

\end{tabular}
\end{adjustbox}
\end{center}
\vspace{-1\baselineskip}
\caption{LMO and YCB average accuracy results.
\label{tab:resultscombined}
}
\vspace{-1.5\baselineskip}
\end{table}

%% file: tabs/e2evs2s_combined.tex
\begin{table}[t]
\begin{center}
\begin{adjustbox}{max width=\columnwidth}
\begin{tabular}{c|c|c|c|c|c}
\cline{2-5}
\multicolumn{1}{c|}{}
& \multicolumn{2}{c|}{mIoU}& \multicolumn{2}{c|}{VCS}\\\hline
Architecture   & train & test&train&test & Loss
\\ \hline \hline
\multirow{2}{*}{Parallel} 
& 85.6 & 63.2 &54.5&52.1
& $\mathcal{L}_{s}$\\\cline{2-6}
 & 91.2 &\underline{76.3}&63.5&\underline{61.7}
 & $\mathcal{L}_{r}\!+\!\mathcal{L}_{bce}$\\
 \cline{1-6}
Cascade & 100 & \textbf{91.7}&95.4&\textbf{93.7}& $\mathcal{L}_{r}$, $\mathcal{L}_{bce}$
\\\hline

\end{tabular}
\end{adjustbox}
\end{center}
\vspace{-1\baselineskip}
\caption{Performance of segmentation  for parallel vs. cascade architectures and varying losses, on LMO} 
 \label{tab:e2evs2scombined}
 \vspace{-1.2\baselineskip}
\end{table}

%% file: tabs/e2evs2s_occLINEMOD.tex
\begin{table}
\begin{center}
\begin{adjustbox}{max width=\columnwidth}
\begin{tabular}{c|c|c|c|c|c|c}
\hline
Architecture   & Loss  & ICP &
ADD(s) & memory & capacity & fps
\\\hline \hline
\multirow{4}{*}{Parallel (4SA)}
& \multirow{2}{*}{$\mathcal{L}_{s}$} & \xmark & 49.9 &\multirow{4}{*}{20.4GB}&\multirow{4}{*}{1.88M}&\multirow{4}{*}{16}\\ \cline{3-4}
&                             & \checkmark &  50.3 &&\\ \cline{2-4}
&  \multirow{-0.5}{*}{$\mathcal{L}_{r}\!+\!\mathcal{L}_{bce}$}    & \xmark & 61.5& &&\\ \cline{3-4}
& & \checkmark   & 61.9 &&\\ \hline
\multirow{2}{*}{Parallel (6SA)} &  & \xmark & 63.2 & \multirow{2}{*}{25.7GB} &\multirow{2}{*}{4.31M}&\multirow{2}{*}{15}\\ \cline{3-4}
&  \multirow{-2}{*}{$\mathcal{L}_{r}\!+\!\mathcal{L}_{bce}$}    & \checkmark &63.7 & &&\\ \hline
\multirow{-0.5}{*}{Cascade}
& \multirow{-0.5}{*}{$\mathcal{L}_{r}$, $\mathcal{L}_{bce}$}  & \xmark          & \underline{73.7}&\multirow{2}{*}{25.1GB} &\multirow{2}{*}{3.76M}&\multirow{2}{*}{15}\\ \cline{3-4}
&  & \checkmark                          & \textbf{74.5}&&\\ \hline

\end{tabular}
\end{adjustbox}
\end{center}
\vspace{-1\baselineskip}
\caption{Pose estimation accuracy (ADD(s)) for parallel vs. cascade architectures and varying losses, on LMO  
\label{tab:e2evs2s}\vspace{+10pt}
}
\vspace{-2\baselineskip}
\end{table}

%% file: tabs/desciptors_combined.tex
\begin{table}[t]
\begin{center}
\begin{adjustbox}{max width=\columnwidth}
\begin{tabular}{c|c|c|c|c}
\cline{3-5}
\multicolumn{2}{c|}{}
&
LMO & \multicolumn{2}{|c}{YCB}\\ 
\hline
\multirow{2}{*}{3D Descriptors} 
& \multirow{2}{*}{ICP}     
& \multirow{2}{*}{ADD(S)}  
& ADD-S   & ADD(S)
\\
& & & AUC & AUC \\ \hline \hline
\multirow{2}{*}{PointNet++~\cite{qi2017pointnet++}}
& \xmark  & 73.7 & 96.3 & 95.7  \\ \cline{2-5}
& \checkmark& \textbf{74.5} & \textbf{96.6} & \textbf{96.0}\\ \hline
\multirow{2}{*}{Point Transformer~\cite{zhao2021point}} & \xmark  & 73.3 & 95.7 & 95.3 \\ \cline{2-5}
 & \checkmark  & \underline{73.8} & \underline{96.0} & \underline{95.6} \\ \hline
\multirow{2}{*}{PPF+knn} & \xmark & 70.8 & 95.5 & 95.0 \\\cline{2-5}
& \checkmark & 71.7 & 95.8 & 95.2 \\\hline
\multirow{2}{*}{FPFH+knn} & \xmark & 70.6 & 94.6 & 93.2 \\\cline{2-5}
& \checkmark & 71.3 & 94.9 & 93.6 \\\hline
\multirow{2}{*}{DGCNN~\cite{wang2019dgcnn}} & \xmark & 70.1 & 92.8 & 90.9 \\\cline{2-5}
& \checkmark & 70.7 & 93.0 & 91.1 \\\hline
\multirow{2}{*}{SubdivNet~\cite{hu2021subdivnet}} & \xmark & 65.4 & 92.0 & 90.4 \\\cline{2-5}
& \checkmark & 66.2 & 92.2 & 90.6 \\\hline

\end{tabular}
\end{adjustbox}
\end{center}
\vspace{-1\baselineskip}
\caption{Average accuracies
of varied 3D descriptors 
\label{tab:descriptors_combined}
}
\vspace{-1.0\baselineskip}
\end{table}

%% file: figs/fig-NoofVotes.tex
\begin{figure}[t]
\begin{center}
\includegraphics[width=0.8\columnwidth]{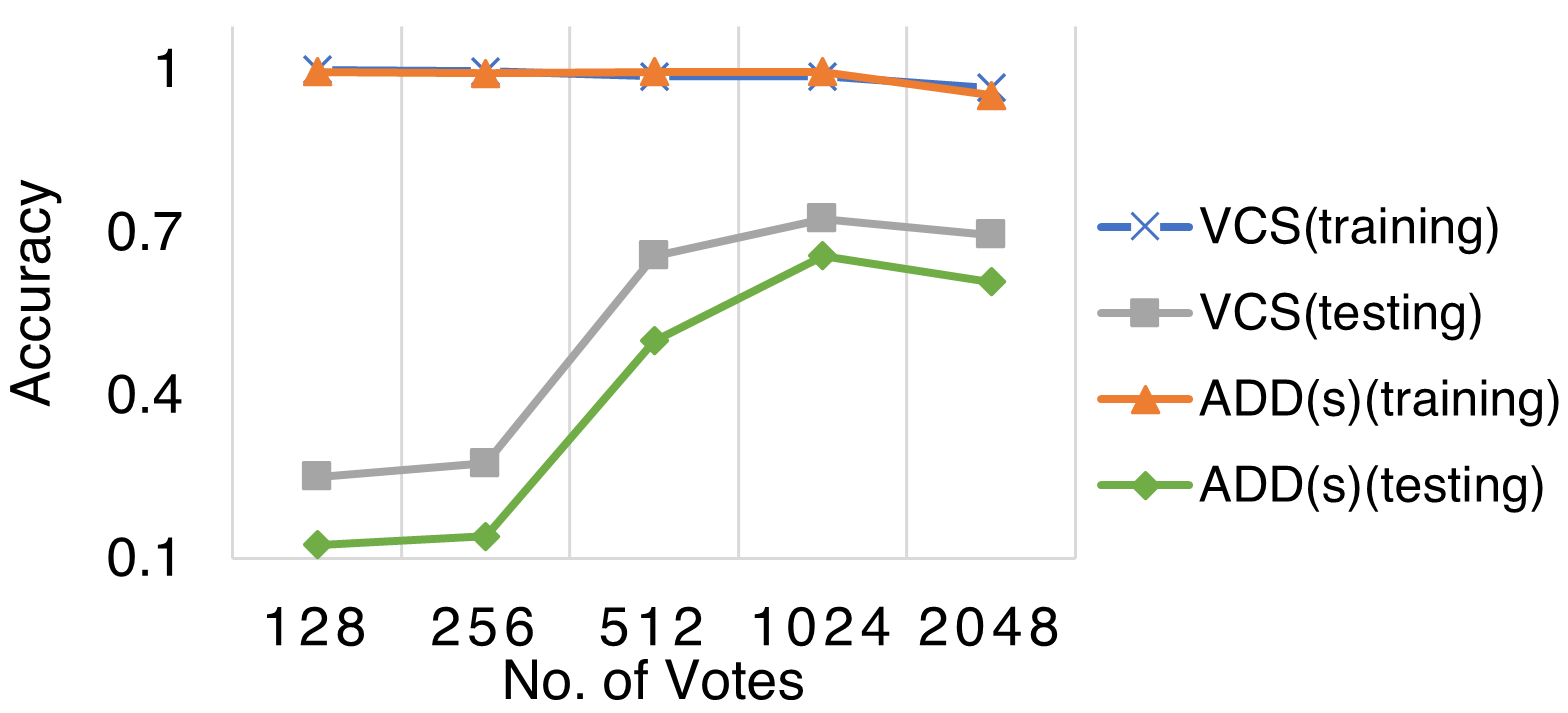}
\vspace{-0.5\baselineskip}
  \caption{Impact of number of votes on accuracy (VCS and ADD(s)), tested on \emph{ape} in LMO. The best is 1024 points }
  \label{fig:NoofVotes}
\end{center}
\vspace{-2\baselineskip}
\end{figure}

%% file: tabs/geoLoss.tex
\begin{table}[t]
\begin{center}
\begin{tabular}{lcc|cc}
\cline{2-5}
& \multicolumn{2}{c|}{$\mathcal{L}_{\epsilon}$} & \multicolumn{2}{c}{$\mathcal{L}_{r}
=\alpha \mathcal{L}_{\epsilon} + \beta \cal{L}_P$} \\\hline\hline
initial learning rate  & $1e^{-3}$    & $1e^{-4}$    & $1e^{-3}$        & $1e^{-4}$       \\
convergence time &       5h       &     8h         &        \textbf{3h}          &       \underline{4h}          \\
VCS        &     $45\%$         &    $87\%$          &      \underline{$90\%$}            &     $\textbf{94\%}$           \\
ADD(s) & 47.5 & 68.2 & \underline{73.2} &$\textbf{74.5}$\\ \hline
\end{tabular}
\end{center}
\vspace{-1\baselineskip}
\caption{Impact of Radial Pair Loss $\cal{L}_P$ on radii regression}
\vspace{-1.5\baselineskip}
\label{tab:geoloss}
\end{table}

%% file: tabs/convergenceTrainableParam.tex
\begin{table}[ht]
\begin{center}
\resizebox{\linewidth}{!}{
\begin{tabular}{c|c|c}
\hline
3D Descriptor   & Time & $\#$ Parameters   \\ \hline\hline
PointNet++ & \textbf{16h}                   &     \textbf{1.8M}                 \\ \hline
Point Transformer~\cite{zhao2021point} & 22h & 4.7M \\\hline
PPF+knn    & 48h                   &     9.7M                \\ \hline
FPFH+knn    & 48h                   &    9.8M                    \\ \hline
DGCNN~\cite{wang2019dgcnn}      & 70h                   &    10.2M            \\ \hline
SubdivNet~\cite{hu2021subdivnet}  & 65h                   &    5.8M             \\ \hline
\end{tabular}
}
\end{center}
\caption{Training time and number of trainable parameters comparison of different 3D feature descriptors}
\label{tab:convergenceAndTParam}
\end{table}

%% file: tabs/occLINEMOD-rev2.tex
\begin{table*}[ht]
\begin{center}
\begin{adjustbox}{max width=\textwidth}
\begin{tabular}{c|c|c|c|c|c|c|c|c|c|c}
\cline{3-11}
\multicolumn{2}{c|}{}
&
\multicolumn{7}{c}{Object} & \multirow{2}{*}{hole-}  
\\
\cline{1-2}
Mode      & Method     & ape  & can  & cat  & driller & duck & eggbox$^{*}$ & glue$^{*}$ & puncher &
Mean
\\\hline \hline
                                & Oberweger~\cite{oberweger2018making}  & 12.1 & 39.9 & 8.2  & 45.2    & 17.2 & 22.1   & 35.8 & 36.0        & 27.1 \\\cline{2-11} 
                                & Hu et al.~\cite{hu2019segmentation}  & 17.6 & 53.9 & 3.3  & 62.4    & 19.2 & 25.9   & 39.6 & 21.3        & 30.4 \\\cline{2-11} 
                                & Pix2Pose~\cite{park2019pix2pose}   & 22.0 & 44.7 & 22.7 & 44.7    & 15.0 & 25.2   & 32.4 & 49.5        & 32.0 \\\cline{2-11} 
                                & DPOD~\cite{zakharov2019dpod}       & -  & -  & -  & -     & -  & -    & -  & -         & 32.8 \\\cline{2-11} 
                                & PVNet~\cite{pvnet}      & 15.8 & 63.3 & 16.7 & 25.2    & 65.7 & 50.2   & 49.6 & 39.7        & 40.8 \\\cline{2-11} 
\multirow{-6}{*}{RGB}
& PPRN~\cite{trabelsi2021ppn}       & -  & -  & -  & -     & -  & -    & -  & -         & 58.4 \\\hline\hline
                                & YOLO6D~\cite{yolo6d}     & -  & -  & -  & -     & -  & -    & -  & -         & 6.4  \\\cline{2-11} 
                                & SSD6D+ref~\cite{kehl2017ssd}  & -  & -  & -  & -     & -  & -    & -  & -         & 27.5 \\\cline{2-11} 
                               
\multirow{-3}{*}{RGB}      
&  PoseCNN~\cite{posecnn}    & 9.6  & 45.2 & 0.9  & 41.4    & 19.6 & 22.0   & 38.5 & 22.1        & 24.9 \\\cline{2-11} 
\multirow{-3}{*}{+D ref}
& DPOD+ref~\cite{zakharov2019dpod}   & -  & -  & -  & -     & -  & -    & -  & -         & 47.3 \\\hline\hline

                                & RCVPose     & 60.3 & 92.5 & 50.2 & 78.2    & 52.1 & 81.2   & 72.1  & 75.2        & 70.2 \\\cline{2-11} 
\multirow{+3}{*}{RGB-D/RGB+3D}  

& RCVPose+ICP & 61.3 & 93 & 51.2 & 78.8  & 53.4 & \textbf{82.3}  & \underline{72.9} & 75.8 & 71.1\\ \cline{2-11} 
& PVN3D~\cite{pvn3d}      & 33.9 & 88.6 & 39.1 & 78.4   & 41.9 & 80.9   & 68.1 & 74.7        & 63.2 \\\cline{2-11} 
& FFB6D~\cite{he2021ffb6d}      & - & - & - & - & - & -   & - & -   & 66.2 \\\cline{2-11} 
& PointVoteNet+ICP\cite{hagelskjaer2020pointvotenet}   & \textbf{70.0} & \underline{95.5} & \textbf{60.8} & \textbf{87.9}   & \textbf{70.7} & 58.7   & 66.9 & \textbf{90.6}        & \textbf{75.1} \\\hline\hline
\multirow{2}{*}{3D/Mesh}
& RCVPose3D & 63   & 96.1 & 57.3 & 81.4  & 54.2 & 81.5  & 75.1 & 80.7 & 73.7\\ \cline{2-11}
& RCVPose3D + ICP & \underline{65.7} & \textbf{96.4} & \underline{58.5} & \underline{81.7}  & \underline{54.7} & \underline{82}    & \textbf{75.4} & \underline{81.2} & \underline{74.5}\\ \hline


\end{tabular}
\end{adjustbox}
\end{center}
\caption{LMO accuracy results:
accuracy of 6DoF PE methods for non-symmetric objects are evaluated with ADD, and that of 
symmetric objects
(annotated with~$^{*}$) are evaluated with ADD-s. 
\label{tab:OccLinemod}
}
\vspace{2\baselineskip}
\end{table*}

%% file: tabs/occlinemod_descriptors-rev2.tex
\begin{table*}[ht]
\begin{center}
\begin{adjustbox}{max width=\textwidth}
\begin{tabular}{c|c|c|c|c|c|c|c|c|c|c}
\cline{3-11}
\multicolumn{2}{c|}{}
&
\multicolumn{7}{c}{Object} & \multirow{2}{*}{hole-}  
& 
\\\cline{1-2}
Descriptors      & ICP     & ape  & can  & cat  & driller & duck & eggbox$^{*}$ & glue$^{*}$ & puncher &
Mean
\\\hline \hline
\multirow{2}{*}{PointNet++~\cite{qi2017pointnet++}}
& \xmark & 63   & 96.1 & 57.3 & 81.4  & 54.2 & 81.5  & 75.1 & 80.7 & 73.7\\ \cline{2-11}
& \checkmark & \textbf{65.7} & \textbf{96.4} & \textbf{58.5} & \textbf{81.7}  & \underline{54.7} & \textbf{82}    & \textbf{75.4} & 81.2 & \textbf{74.5}\\ \hline
\multirow{2}{*}{Point Transformer~\cite{zhao2021point}} & \xmark & 59.2 & 93.4 & 56.8 & 80.8 & 58.2 & 79.6 & 73.2 & 85.4 & 73.3 \\ \cline{2-11}
 & \checkmark & \underline{59.9}  & \underline{93.7} & \underline{57.2} & \underline{81.1} & \textbf{58.6} & \underline{80.1} & 73.6 & \underline{85.8} & \underline{73.8} \\ \hline
\multirow{2}{*}{PPF+knn} & \xmark & 55.7 & 88.6 & 51.2 &  78.2 & 54.1 & 79.4  & 73.8 & 85.7 & 70.8 \\\cline{2-11}
& \checkmark & 56.8 & 89.4 & 51.6 & 78.8 & 54.8 & 80.2  & \underline{74.3} & \textbf{87.2} & 71.7 \\\hline
\multirow{2}{*}{FPFH+knn} & \xmark & 55.4 & 89.2 & 50.2 & 77.8 & 52.6 & 80.9  & 72.4 & 86.3 & 70.6 \\\cline{2-11}
& \checkmark & 55.7 & 90.2 & 50.8 & 78.2  & 53.8 & 81.4 & 73.2 & \textbf{87.2} & 71.3 \\\hline
\multirow{2}{*}{DGCNN\cite{wang2019dgcnn}} & \xmark & 58 & 87.2 & 53.4 & 77.8  & 51.9 & 78.7  & 71.2 & 82.3 & 70.1\\\cline{2-11}
& \checkmark & 59.2 & 88.6 & 54.1 & 78.2  & 52.1 & 78.7  & 71.7 & 82.7 & 70.7 \\\hline
\multirow{2}{*}{SubdivNet\cite{hu2021subdivnet}} & \xmark & 42.3 & 83.2 & 49.2 &  77.8 & 48.7 & 77.9  & 69.2 & 75.2 & 65.4\\\cline{2-11}
& \checkmark & 45.2 & 83.7 & 49.7 & 78.2  & 49.2 & 78.2  & 70.1 & 75.2 & 66.2\\\hline

\end{tabular}
\end{adjustbox}
\end{center}
\caption{LMO accuracy results for different descriptors:
accuracy of RCVPose3D for non-symmetric objects is evaluated with ADD, and for
symmetric objects
(annotated with~$^{*}$) is evaluated with ADD-s.
\label{tab:OccLinemodDescriptors}
}
 \vspace{-0.7\baselineskip}
\end{table*}

%% file: tabs/YCB.tex
\begin{sidewaystable*}[ht]
\begin{center}
\begin{adjustbox}{max width=\textwidth}
\begin{tabular}{|ccc|c|c|c|c|c|c|c|c|c|c|c|c|c|c|c|c|c|c|c|c|c|c|}
\multicolumn{1}{c}{}&&&
\begin{rotate}{60}
\hspace{-0.65 cm} 002 master
\end{rotate}
&
\begin{rotate}{60}
\hspace{-0.65 cm} 003 cracker
\end{rotate}
&
\begin{rotate}{60}
\hspace{-0.65 cm} 004 sugar
\end{rotate}
&
\begin{rotate}{60}
\hspace{-0.65 cm} 005 tomato  
\end{rotate}
&
\begin{rotate}{60}
\hspace{-0.65 cm} 006 mustard  
\end{rotate}
&
\begin{rotate}{60}
\hspace{-0.65 cm} 007 tuna fish  
\end{rotate}
&
\begin{rotate}{60}
\hspace{-0.65 cm} 008 pudding 
\end{rotate}
&
\begin{rotate}{60}
\hspace{-0.65 cm} 009 gelatin
\end{rotate}
&
\begin{rotate}{60}
\hspace{-0.65 cm} 010 potted  
\end{rotate}
&
\begin{rotate}{60}
\hspace{-0.65 cm} 011 banana  
\end{rotate}
&
\begin{rotate}{60}
\hspace{-0.65 cm} 019 pitcher 
\end{rotate}
&
\begin{rotate}{60}
\hspace{-0.65 cm} 021 bleach 
\end{rotate}
&
\begin{rotate}{60}
\hspace{-0.65 cm} 024 bowl$^{*}$ 
\end{rotate}
&
\begin{rotate}{60}
\hspace{-0.65 cm} 025 mug 
\end{rotate}
&
\begin{rotate}{60}
\hspace{-0.65 cm} 035 power 
\end{rotate}
&
\begin{rotate}{60}
\hspace{-0.65 cm} 036 wood 
\end{rotate}
&
\begin{rotate}{ 60}
\hspace{-0.65 cm} 037 scissors 
\end{rotate}
&
\begin{rotate}{60}
\hspace{-0.65 cm} 040 large 
\end{rotate}
&
\begin{rotate}{60}
\hspace{-0.65 cm} 051 large 
\end{rotate}
&
\begin{rotate}{60}
\hspace{-0.65 cm} 052 extra large$^{*}$ 
\end{rotate}
&
\begin{rotate}{60}
\hspace{-0.65 cm} 061 foam 
\end{rotate}
&\\

\cline{1-3} 
\multicolumn{1}{|c}{Refine} & \multicolumn{1}{|c}{Metric}  & \multicolumn{1}{|c|}{Method}                         & 
\begin{rotate}{60}
\hspace{0.8 cm} {chef can}
\end{rotate}

& 
\begin{rotate}{60}
\hspace{0.8 cm}
{box} 
\end{rotate}
&
\begin{rotate}{60}
\hspace{0.8 cm}{box} 
\end{rotate}
&
\begin{rotate}{60}
\hspace{0.8 cm}{soup can} 
\end{rotate}
&
\begin{rotate}{60}
\hspace{0.8 cm}{bottle} 
\end{rotate}
&
\begin{rotate}{60}
\hspace{0.8 cm}{can} 
\end{rotate}
&
\begin{rotate}{60}
\hspace{0.8 cm}{box} 
\end{rotate}
&
\begin{rotate}{60}
\hspace{0.8 cm}{box} 
\end{rotate}
&
\begin{rotate}{60}
\hspace{0.8 cm}{meat can} 
\end{rotate}
&
&
\begin{rotate}{60}
\hspace{0.8 cm}{base} 
\end{rotate}
&
\begin{rotate}{60}
\hspace{0.8 cm}{cleanser} 
\end{rotate}
&
&
&
\begin{rotate}{60}
\hspace{0.8 cm}{drill} 
\end{rotate}
&
\begin{rotate}{60}
\hspace{0.8 cm}{block$^{*}$} 
\end{rotate}
&
&
\begin{rotate}{60}
\hspace{0.8 cm}{marker} 
\end{rotate}
&
\begin{rotate}{60}
\hspace{0.8 cm}{clamp$^{*}$} 
\end{rotate}
&
\begin{rotate}{60}
\hspace{0.8 cm}{clamp}
\end{rotate}
&
\begin{rotate}{60}
\hspace{0.8 cm}{brick$^{*}$} 
\end{rotate}
&
{Mean}   
\\ \hline \hline
\multicolumn{1}{|c|}{{}}       & \multicolumn{1}{c|}{}                          & {PoseCNN~\cite{posecnn} }       & {83.9}                & {76.9}            & {84.2}          & {81.0}                & {90.4}               & {88.0}              & {79.1}            & {87.2}            & {78.5}                & {86.0}       & {77.0}             & {71.6}                & {69.6}     & {78.2}    & {72.7}            & {64.3}           & {56.9}          & {71.7}             & {50.2}            & { 44.1}                  & {88.0}           & {75.8}          \\ \cline{3-25} 
& \multicolumn{1}{|c|}{}                          & {BaseNet~\cite{gao20206d}}    & 95.4 & 93 & 98.5 & 96.5 & 97.7 & 97.7 & 97.3 & \textbf{99} & 95.7 & 97.7 & \underline{97.9} & 97.4 & \textbf{97.7} & 97.8 & \textbf{97.7} & \textbf{94.9} & 91.3 & 98 & 77.4 & 66.4 & \textbf{98} & 94.4      \\ \cline{3-25}   
\multicolumn{1}{|c|}{{}}       & \multicolumn{1}{c|}{ADD-S}                       & {DF(per-pixel)\cite{wang2019densefusion}} & {95.3}                & {92.5}            & {95.1}          & {93.8}                & {95.8}               & {95.7}              & {94.3}            & {97.2}            & {89.3}                & {90.0}       & {93.6}             & {94.4}                & {86.0}     & {95.3}    & {92.1}            & {89.5}           & {90.1}          & {95.1}             & {71.5}            & {70.2}                  & {92.2}           & {91.2}          \\ \cline{3-25} 
\multicolumn{1}{|c|}{{}}       & \multicolumn{1}{c|}{AUC}                              & {PVN3D\cite{pvn3d}}         & { \underline{96.0}}       & { 96.1}            & \underline{ 97.4}          & { 96.2}                & { 97.5}               & 96                                       & \underline{ 97.1}            & { 97.7}            & { 93.3}                & 96.6                              & { 97.4}    & { 96.0}                & {90.2}     & { 97.6}    & { {96.7}}   & {{90.4}}  & { \underline{96.7}} & { \textbf{96.7}}    & { 93.6}            & {88.4}                  & { {96.8}}  & { \textit{95.5}} \\ \cline{3-25} 

& \multicolumn{1}{|c|}{}                          & {FFB6D\cite{he2021ffb6d}}         & \textbf{96.3}                & { 96.3}            & \textbf{97.6}                                 & {95.6}                & \underline{97.8}                                      & \underline{96.8}              & \underline{ 97.1}            & \underline{98.1}                                   & { 94.7}                & { 97.2}       & {97.6}                           & { 96.8}                & \underline{96.3}     & { 97.3}    & \underline{ 97.2}            & {{92.6}}  & \textbf{97.7}          & \underline{ 96.6}             & \textbf{ 96.8}            & \textbf{96.0}                  & { \textbf{97.3}}  & \textbf{ 96.6}          \\ \cline{3-25} 

\multicolumn{1}{|c|}{{}}       & \multicolumn{1}{c|}{}     & RCVPose\cite{wu2021vote}                              & 95.7                                       & \textbf{97.2}                          & \textbf{97.6}                        & \textbf{98.2}                              & \textbf{97.9}                           & \textbf{98.2}                            & \textbf{97.7}                          & 97.7                                   & \underline{97.9}                            & \underline{97.9}                    & 96.2                                    & \textbf{99.2}                              & 95.2                   & \underline{98.4}                  & 96.2                                   & 89.1                                  & 96.2                                 & 95.9                                    & \underline{95.2}   & \underline{94.7}      & 95.7                                  & \textbf{96.6}                    \\ \cline{3-25} 
& \multicolumn{1}{|c|}{} & RCVPose3D &   92.1 & \underline{96.9} & \underline{97.4} & \underline{96.9} & 97.6 & 96.7 & \textbf{97.7} & 97.8 & \textbf{98.7} & \textbf{98.2} & \textbf{98.4} & \underline{98.6} & 92.2 & \textbf{98.6} & 97.1 & \underline{93.2} & \textbf{97.7} & \textbf{96.7} & 92   & 93.4 & 95   & \underline{96.3} \\\cline{2-25} 

\multicolumn{1}{|c|}{{}}       & \multicolumn{1}{c|}{}                          & {PoseCNN~\cite{posecnn} }       & {50.2}                & {53.1}            & {68.4}          & {66.2}                & {81.0}               & {70.7}              & {62.7}            & {75.2}            & {\textit{59.5}}       & {72.3}       & {53.3}             & {50.3}                & {69.6}     & {58.5}    & {55.3}            & {64.3}           & {35.8}          & {58.3}             & {50.2}            & { 44.1}                  & {88.0}           & {59.9}          \\ \cline{3-25}
\multicolumn{1}{|c|}{\multirow{-4}{*}{No}}       
& \multicolumn{1}{c|}{}                          & {DF(per-pixel)\cite{wang2019densefusion}} & {70.7}                & {86.9}            & {90.8}          & {84.7}                & {90.9}               & {79.6}              & {89.3}            & {95.8}            & {79.6}                & {76.7}       & {87.1}             & {87.5}                & {86.0}     & {83.8}    & {83.7}            & {89.5}           & {77.4}          & {89.1}             & {71.5}            & {70.2}                  & {92.2}           & {82.9}          \\ \cline{3-25} 

& \multicolumn{1}{|c|}{ADD(S)}                          & {PVN3D\cite{pvn3d}}         & { 80.5}                & { 94.8}            & 96.3                                 & {88.5}                & 96.2                                      & {89.3}              & { 95.7}            & 96.1                                   & { 88.6}                & { 93.7}       & 96.5                           & { 93.2}                & {90.2}     & { 95.4}    & { 95.1}            & {90.4}  & { 92.7}          & { 91.8}             & { 93.6}            & {88.4}                  & \underline{ 96.8}  & { 91.8}          \\ \cline{3-25} 
& \multicolumn{1}{|c|}{AUC}                          & {FFB6D\cite{he2021ffb6d}}         & { 80.6}                & { 94.6}            & 96.6                                 & {89.6}                & \underline{97}                                      & {88.9}              & { 94.6}            & 96.9                                   & { 88.1}                & { 94.9}       & \underline{96.9}                           & { 94.8}                & \underline{96.3}     & { 94.2}    & \underline{ 95.9}            & {\underline{92.6}}  & \underline{ 95.7}          & { 89.1}             & \textbf{ 96.8}            & {\textbf{96.0}}                  & { \textbf{97.3}}  & { 92.7}          \\ \cline{3-25} 

& \multicolumn{1}{|c|}{}                          & {Basenet\cite{gao20206d}}         & { 46.9}                & { 76.7}            & \textbf{97.5}                                 & {72.7}                & 79.2                                      & {72}              & { 94.4}            & \textbf{98.6}                                   & \underline{ 90.6}                & { 95.1}       & {96.1}                           & { 95.4}                & {83.9}     & { 93.9}    & { 94.9}            & {{90}}  & { 75.8}          & { 92.2}             & { 68.5}            & {25.3}                  & { {92.9}}  & { 82.5}          \\ \cline{3-25}    

& \multicolumn{1}{|c|}{} & RCVPose                               & \textbf{93.6}                              & \underline{95.7}                        & \underline{97.2}                        & \textbf{94.7}                            & \textbf{97.2}                             & \textbf{96.4}                            & \textbf{97.1}                          & 96.5                          & 90.2                             & \underline{96.7}                     & 95.7                                    & \underline{97.8}                              & \textbf{95.2}                   & \underline{96.3}                  & 95.4                         & 89.1                                  & \underline{94.7}                        & \underline{92.4}                          & \textbf{95.2} & 94.7     & 95.7                                  & \underline{95.2}                   \\\cline{3-25}
& \multicolumn{1}{|c|}{} & RCVPose3D  &   \underline{91.7} & \textbf{96.2} & 95.4 & \underline{94.3} & 95.6 & \underline{96.2} & \underline{96.7} & \underline{97}  & \textbf{98.2} & \textbf{97.9} & \textbf{97.7} & \textbf{98.2} & 92.2 & \textbf{98.6} & \textbf{96.7} & \textbf{93.2} & \textbf{97.2} & \textbf{96.4} & 92   & 93.4 & 95   & \textbf{95.7}  \\

\hline \hline
\multicolumn{1}{|c|}{{}}       & \multicolumn{1}{c|}{}                          & {PoseCNN~\cite{posecnn} +ICP}   & \underline{ 95.8}                & { 92.7}            & { 98.2}          & { 94.5}                & { \textbf{98.6}}               & \underline{ 97.1}              & { 97.9}            & \textbf{ 98.8}            & { 92.7}                & { 97.1}       & \underline{ 97.8}             & { 96.9}                & {81.0}     & { 94.9}    & \textbf{ 98.2}            & {87.6}           & { 91.7}          & { 97.2}             & {75.2}            & {64.4}                  & \underline{ 97.2}           & { 93.0}          \\ \cline{3-25} 
\multicolumn{1}{|c|}{{}}       & \multicolumn{1}{c|}{}                          & {DF(iterative)\cite{wang2019densefusion}} & { 96.4}                & { 95.8}            & { 97.6}          & { 94.5}                & { 97.3}               & \underline{ 97.1}              & { 96.0}            & { 98.0}            & { 90.7}                & { 96.2}       & { 97.5}             & { 95.9}                & {89.5}     & { 96.7}    & { 96.0}            & \underline{ 92.8}           & { 92.0}          & \underline{ 97.6}             & {72.5}            & { 69.9}                  & {92.0}           & { 93.2}          \\ \cline{3-25} 
\multicolumn{1}{|c|}{{}}       & \multicolumn{1}{c|}{ADD-S}                          & {PVN3D\cite{pvn3d}+ICP}     & { 95.2}                & {94.4}            & \textbf{97.9}          & { 95.9}                & \underline{98.3}      & {96.7}              & { \textbf{98.2}}   & { \textbf{98.8}}   & { 93.8}                & { 98.2}       & {97.6}    & { 97.2}                & \underline{92.8}     & { 97.7}    & {97. 1}  & {91.1}  & {95.0}          & \textbf{ 98.1}             & \underline{ 95.6}            & { 90.5}                  & { \textbf{98.2}}  & { 96.1}          \\ \cline{3-25} 
\multicolumn{1}{|c|}{{}}       & \multicolumn{1}{c|}{AUC}     & RCVPose+ICP                           & \textbf{96.2}                              & \textbf{97.9}                          & \textbf{97.9}                                 & \textbf{99}                                & 98.2                                      & \textbf{98.6}                            & \underline{98.1}                                   & \underline{98.4}                                   & \underline{98.4}                              & \underline{98.3}                    & 97.2                                    & \textbf{99.6}                              & \textbf{96.9}                   & \textbf{98.7}                  & 96.4                                   & 90.7                                  & \underline{96.4}                        & 96.6                          & \textbf{96.2}   & \textbf{95.1}      & 96.6                                  & \textbf{97.2}                    \\ \cline{3-25} 

& \multicolumn{1}{|c|}{} & RCVPose3D & 92.4 & \underline{97.2} & \underline{97.7} & \underline{97.2} & 97.9 & 97   & 98   & 98   & \textbf{99}   & \textbf{98.5} & \textbf{98.7} & \underline{98.8} & 92.4 & \underline{98.6} & \underline{97.4} & \textbf{93.7} & \textbf{97.9} & 97   & 92.3 & \underline{93.7} & 95.2 & \underline{96.6}\\\cline{2-25}

\multicolumn{1}{|c|}{{}}       & \multicolumn{1}{c|}{}                          & {PoseCNN~\cite{posecnn} +ICP}   & { 68.1}                & {83.4}            & { 97.1}          & {81.8}                & { 98.0}               & {83.9}              & { 96.6}            & \underline{98.1}            & {83.5}                & { 91.9}       & \underline{ 96.9}             & { 92.5}                & {81.0}     & {81.1}    & { 97.7}            & {87.6}           & {78.4}          & {85.3}             & {75.2}            & { 64.4}                  & \underline{ 97.2}           & {85.4}          \\ \cline{3-25} 
\multicolumn{1}{|c|}{\multirow{-4}{*}{Yes}}       & \multicolumn{1}{c|}{}                          & {DF(iterative)\cite{wang2019densefusion}} & {73.2}                & { 94.1}            & { 96.5}          & {85.5}                & { 94.7}               & {81.9}              & { 93.3}            & { 96.7}            & {83.6}                & {83.3}       & \underline{ 96.9}             & {89.9}                & {89.5}     & {88.9}    & { 92.7}            & \underline{ 92.8}           & {77.9}          & { 93.0}             & {72.5}            & {69.9}                  & { 92.0}           & {86.1}          \\ \cline{3-25} 
& \multicolumn{1}{|c|}{ADD(S)}                          & {PVN3D\cite{pvn3d}+ICP}     & {79.3}                & { 91.5}            & \underline{ 96.9}          & { 89.0}                & { \textbf{97.9}}      & { 90.7}              & \underline{ 97.1}            & { \textbf{98.3}}   & {87.9}                & { 96.0}       & \underline{ 96.9}    & \underline{ 95.9}                & \underline{ 92.8}     & { 96.0}    & { 95.7}            & { 91.1}           & {87.2}          & { 91.6}             & \underline{ 95.6}            & { 90.5}                  & { \textbf{98.2}}  & { 92.3}          \\ \cline{3-25} 
& \multicolumn{1}{|c|}{AUC} & RCVPose+ICP                           & \textbf{94.7}                              & \underline{96.4}                       & \textbf{97.6}                        & \textbf{95.4}                              & \underline{97.7}                                     & \textbf{96.7}                            & \textbf{97.4}                          & 97.9                                 & \underline{92.6}                             & \underline{97.2}                    & 96.7                                    & \textbf{98.4}                              & \textbf{95.3}                   & \underline{97.1}               & \underline{96.9}                          & 90.7                         & \underline{94.9}                       & \underline{93.2}                           & \textbf{96.2}    & \textbf{95.1}      & 96.6                                  & \underline{95.9}                    \\ \cline{3-25}

& \multicolumn{1}{|c|}{} & RCVPose3D  & \underline{92}   & \textbf{96.8} & 95.6 & \underline{94.6} & 95.9 & \underline{96.4} & 97   & 97.3 & \textbf{98.4} & \textbf{98.2} & \textbf{98}   & \textbf{98.4} & 92.5 & \textbf{98.8} & \textbf{97}   & \textbf{93.7} & \textbf{97.2} & \textbf{96.7} & 92.3 & \underline{93.7} & 95.2 & \textbf{96.0}\\\hline
\end{tabular}
\end{adjustbox}
\caption{YCB accuracy results: AUC~\cite{posecnn} and ADD(s)~\cite{hinterstoisser2012model} are used as follows: 6DoF PE methods are evaluated with ADD for non-symmetric objects, and are evaluated with ADD-S for
symmetric objects
(annotated with~$^{*}$) while computing for ADD(S) AUC. 
\label{tab:YCBVideoFull}
}
\end{center}

\end{sidewaystable*}

%% file: tabs/ycb_descriptors.tex
\begin{sidewaystable*}[]
\begin{center}
\begin{adjustbox}{max width=\textwidth}
\begin{tabular}{|ccc|c|c|c|c|c|c|c|c|c|c|c|c|c|c|c|c|c|c|c|c|c|c|}
\multicolumn{1}{c}{}&&&
\begin{rotate}{60}
\hspace{-0.65 cm} 002 master
\end{rotate}
&
\begin{rotate}{60}
\hspace{-0.65 cm} 003 cracker
\end{rotate}
&
\begin{rotate}{60}
\hspace{-0.65 cm} 004 sugar
\end{rotate}
&
\begin{rotate}{60}
\hspace{-0.65 cm} 005 tomato  
\end{rotate}
&
\begin{rotate}{60}
\hspace{-0.65 cm} 006 mustard  
\end{rotate}
&
\begin{rotate}{60}
\hspace{-0.65 cm} 007 tuna fish  
\end{rotate}
&
\begin{rotate}{60}
\hspace{-0.65 cm} 008 pudding 
\end{rotate}
&
\begin{rotate}{60}
\hspace{-0.65 cm} 009 gelatin
\end{rotate}
&
\begin{rotate}{60}
\hspace{-0.65 cm} 010 potted  
\end{rotate}
&
\begin{rotate}{60}
\hspace{-0.65 cm} 011 banana  
\end{rotate}
&
\begin{rotate}{60}
\hspace{-0.65 cm} 019 pitcher 
\end{rotate}
&
\begin{rotate}{60}
\hspace{-0.65 cm} 021 bleach 
\end{rotate}
&
\begin{rotate}{60}
\hspace{-0.65 cm} 024 bowl$^{*}$ 
\end{rotate}
&
\begin{rotate}{60}
\hspace{-0.65 cm} 025 mug 
\end{rotate}
&
\begin{rotate}{60}
\hspace{-0.65 cm} 035 power 
\end{rotate}
&
\begin{rotate}{60}
\hspace{-0.65 cm} 036 wood 
\end{rotate}
&
\begin{rotate}{ 60}
\hspace{-0.65 cm} 037 scissors 
\end{rotate}
&
\begin{rotate}{60}
\hspace{-0.65 cm} 040 large 
\end{rotate}
&
\begin{rotate}{60}
\hspace{-0.65 cm} 051 large 
\end{rotate}
&
\begin{rotate}{60}
\hspace{-0.65 cm} 052 extra large$^{*}$ 
\end{rotate}
&
\begin{rotate}{60}
\hspace{-0.65 cm} 061 foam 
\end{rotate}
&\\

\cline{1-3} 
\multicolumn{1}{|c}{Refine} & \multicolumn{1}{|c}{Metric}  & \multicolumn{1}{|c|}{Method}                         & 
\begin{rotate}{60}
\hspace{0.8 cm} {chef can}
\end{rotate}

& 
\begin{rotate}{60}
\hspace{0.8 cm}
{box} 
\end{rotate}
&
\begin{rotate}{60}
\hspace{0.8 cm}{box} 
\end{rotate}
&
\begin{rotate}{60}
\hspace{0.8 cm}{soup can} 
\end{rotate}
&
\begin{rotate}{60}
\hspace{0.8 cm}{bottle} 
\end{rotate}
&
\begin{rotate}{60}
\hspace{0.8 cm}{can} 
\end{rotate}
&
\begin{rotate}{60}
\hspace{0.8 cm}{box} 
\end{rotate}
&
\begin{rotate}{60}
\hspace{0.8 cm}{box} 
\end{rotate}
&
\begin{rotate}{60}
\hspace{0.8 cm}{meat can} 
\end{rotate}
&
&
\begin{rotate}{60}
\hspace{0.8 cm}{base} 
\end{rotate}
&
\begin{rotate}{60}
\hspace{0.8 cm}{cleanser} 
\end{rotate}
&
&
&
\begin{rotate}{60}
\hspace{0.8 cm}{drill} 
\end{rotate}
&
\begin{rotate}{60}
\hspace{0.8 cm}{block$^{*}$} 
\end{rotate}
&
&
\begin{rotate}{60}
\hspace{0.8 cm}{marker} 
\end{rotate}
&
\begin{rotate}{60}
\hspace{0.8 cm}{clamp$^{*}$} 
\end{rotate}
&
\begin{rotate}{60}
\hspace{0.8 cm}{clamp}
\end{rotate}
&
\begin{rotate}{60}
\hspace{0.8 cm}{brick$^{*}$} 
\end{rotate}
&
{Mean}   
\\ \hline \hline
& \multicolumn{1}{|c|}{} & PointNet++~\cite{qi2017pointnet++}                    & \underline{92.1}          & \textbf{96.9} & \underline{97.4}          & \textbf{96.9} & \underline{97.6}          & \textbf{96.7} & \textbf{97.7} & \textbf{97.8} & \textbf{98.7} & \textbf{98.2} & \textbf{98.4} & \textbf{98.6} & \textbf{92.2} & \textbf{98.6} & \textbf{97.1} & \underline{93.2}          & \textbf{97.7} & \underline{96.7}                  & \underline{92}            & \textbf{93.4} & \textbf{95}   & \textbf{96.3} \\\cline{3-25} 
& \multicolumn{1}{|c|}{} & PPF+knn                                               & \textbf{92.3} & \underline{96.2}          & \textbf{97.6} & 93.2          & \textbf{97.7} & 93.2          & \underline{97.6}          & 96.9          & \underline{97.9}          & \underline{97.4}          & \underline{97.7}          & \underline{98.2}          & \textbf{92.2} & \underline{97.7}          & \underline{96.9}          & \underline{93.2}          & \underline{96.7}          & \underline{93.2}                  & \textbf{92.3} & \underline{93.2}          & \underline{94.7}          & \underline{95.5}\\\cline{3-25}           
& \multicolumn{1}{|c|}{} & FPFH+knn                                              & 90.2          & 95.6          & 95.1          & 91.7          & 94.9          & 94.3          & 95.7          & \underline{97.2}          & 96.2          & 97.1          & 96.2          & 97.7          & 87.9          & 97.3          & 96.1          & \textbf{93.5} & 96            & \textbf{96.9}         & 91.2          & 92.2          & 93.4          & 94.6 \\\cline{3-25}           
& \multicolumn{1}{|c|}{\multirow{-4}{*}{ADD-S}} & DGCNN~\cite{wang2019dgcnn}     & 87.2          & 94.7          & 95.2          & \underline{93.7}          & 90.4          & \underline{94.5}          & 93.2          & 96.7          & 93.2          & 96            & 95.1          & 93.2          & 84.3          & 94.2          & 95.4          & 91.2          & 95.1          & 94.2                  & 89.6          & 90.9          & 91.1          & 92.8  \\\cline{3-25}         
& \multicolumn{1}{|c|}{\multirow{-4}{*}{AUC}} & SubdivNet~\cite{hu2021subdivnet} & 89.1          & 92.6          & 93.3          & 90.2          & 91.2          & 92.1          & 94.1          & 95.9          & 93.6          & 96.3          & 92.4          & 91.7          & 83.8          & 94.4          & 93.7          & 91.4          & 93.4          & 91.7                  & 90            & 88.7          & 91.4          & 92.0   \\\cline{2-25}

\multirow{-1.5}{*}{No}& \multicolumn{1}{|c|}{} & PointNet++~\cite{qi2017pointnet++}                          & \textbf{91.7} & \textbf{96.2} & \textbf{95.4} & \textbf{94.3} & \textbf{95.6} & \textbf{96.2} & \textbf{96.7} & \textbf{97}   & \textbf{98.2} & \textbf{97.9} & \textbf{97.7} & \textbf{98.2} & \textbf{92.2} & \textbf{98.6} & \textbf{96.7} & \underline{93.2} & \textbf{97.2} & \textbf{96.4} & \underline{92}            & \textbf{93.4} & \textbf{95}   & \textbf{95.7}  \\\cline{3-25}
& \multicolumn{1}{|c|}{} &  PPF+knn                                               & \textbf{91.7} & \underline{95.9}          & \underline{93.4}          & \underline{92.8}          & \underline{94.3}          & \underline{95.1}          & \underline{95.3}          & 95.4          & \underline{97.1}          & \underline{96.4}          & \underline{96.2}          & \underline{97.4}         & \textbf{92.2} & \textbf{98.6} & \underline{96.1}         & \underline{93.2} & \underline{97}            & \underline{95.7}          & \textbf{92.3} & \underline{93.2}          & \underline{94.7} & \underline{95.0} \\\cline{3-25}
& \multicolumn{1}{|c|}{} &  FPFH+knn                                              & 88.7          & 94.8          & \underline{93.7}          & 89.6          & 93.2          & 92.3          & 94.3          & \underline{95.6}          & 94.4          & 95.1          & 93.6          & 96.1          & \underline{87.9}        & \underline{95.7}          & 93.1          & \textbf{93.5}          & 94.3          & 95.2          & 91.2          & 92.2          & 93.4 & 93.2 \\\cline{3-25}  
& \multicolumn{1}{|c|}{\multirow{-4}{*}{ADD(S)}} &  DGCNN~\cite{wang2019dgcnn}    & 86.9          & 93.2          & 91.7          & 90.2          & 88.6          & 94.2          & 91.7          & 93.3          & 90.1          & 94.2          & 91.5          & 90.7          & 80.2          & 92.7          & 92.6          & 88.7          & 92.3          & 93.7          & 89.6          & 90.9          & 91.1 & 90.9  \\\cline{3-25}
& \multicolumn{1}{|c|}{\multirow{-4}{*}{AUC}} &  SubdivNet~\cite{hu2021subdivnet} & \underline{88.8}          & 91.4          & 90            & 89.4          & 90.2          & 91.3          & 92.4          & 94.1          & 91.2          & 93.9          & 91.1          & 88.6          & 80            & 93.1          & 91.5          & 89.2          & 91.7          & 90.9          & 90            & 88.7          & 91.4 & 90.4   \\
\hline \hline

& \multicolumn{1}{|c|}{}                        & PointNet++~\cite{qi2017pointnet++}                         &\underline{92.4} & \textbf{97.2} & \underline{97.7}          & \textbf{97.2} & \underline{97.9}        & \textbf{97}   & \textbf{98}   & \textbf{98}   & \textbf{99}   & \textbf{98.5} & \textbf{98.7} & \textbf{98.8} & \textbf{92.4}          & \textbf{98.6} & \textbf{97.4} & \textbf{93.7} & \textbf{97.9} & \underline{97}            & \underline{92.3}          & \underline{93.7} & \textbf{95.2} & \textbf{96.6}\\\cline{3-25}
& \multicolumn{1}{|c|}{}                        &  PPF+knn                        & \textbf{92.6}         & \underline{96.4}          & \textbf{98.2} & \underline{93.9}         & \textbf{98} & 93.9          & \underline{97.9}          & 97.1          & \underline{98.2}          & \underline{97.6}          & \underline{97.9}          & \underline{98.6}          & \textbf{92.4}          & \underline{97.9}          & \underline{97.2}          & 93.5          & \underline{97}            & 93.4          & \textbf{92.6} & \underline{93.5}          & \underline{95}            & \underline{95.8}   \\\cline{3-25}
& \multicolumn{1}{|c|}{}                        &  FPFH+knn                       & 90.5         & 96            & 95.6          & 92            & 95.2        & \underline{94.7}          & 95.9          & \underline{97.4}          & 96.5          & 97.3          & 96.5          & 97.9          & \underline{87.9}                   & 97.5          & 96.4          & \textbf{93.9}          & 96.4          & \textbf{97.1} & 91.4          & \underline{93.5}          & 93.7          & 94.9    \\\cline{3-25}
& \multicolumn{1}{|c|}{\multirow{-3}{*}{ADD-S}} &  DGCNN~\cite{wang2019dgcnn}     & 87.5         & 95            & 95.4          & 94            & 90.6        & \underline{94.7}          & 93.5          & 96.9          & 93.4          & 96.3          & 95.4          & 93.5          & 84.3                   & 94.4          & 95.7          & 91.2          & 95.3          & 94.4          & 89.8          & 91.1          & 91.3          & 93.0     \\\cline{3-25}
& \multicolumn{1}{|c|}{\multirow{-3}{*}{AUC}} &  SubdivNet~\cite{hu2021subdivnet} & 89.3         & 92.9          & 93.5          & 90.4          & 91.5        & 92.4          & 94.4          & 96.1          & 93.9          & 96.5          & 92.6          & 92            & 83.8                   & 94.7          & 94            & 91.4          & 93.6          & 92            & 90.2          & 89            & 91.6          & 92.2    \\\cline{2-25}

\multirow{-1.5}{*}{Yes}& \multicolumn{1}{|c|}{}  &  PointNet++~\cite{qi2017pointnet++}                            & \textbf{92}   & \textbf{96.8} & \textbf{95.6} & \textbf{94.6} & \textbf{95.9} & \textbf{96.4} & \textbf{97}   & \textbf{97.3} & \textbf{98.4} & \textbf{98.2} & \textbf{98}   & \textbf{98.4} & \textbf{92.5} & \underline{98.8}          & \textbf{97}   & \textbf{93.7} & \textbf{97.2} & \textbf{96.7} & \underline{92.3}          & \textbf{93.7} & \textbf{95.2} & \textbf{96.0}\\\cline{3-25}
& \multicolumn{1}{|c|}{}                         &  PPF+knn                            & \textbf{92}   & \underline{96.2}          & 93.7          & \underline{93.1}          & \underline{94.5}          & \underline{95.4}          & \underline{95.9}          & 95.6          & \underline{97.3}          & \underline{96.6}          & \underline{96.4}          & \underline{97.6}          & \underline{92.4}          & \textbf{98.9} & \underline{96.4}          & \underline{93.5}          & \underline{97.1}          & \underline{96}            & \textbf{92.6} & 93.5          & \underline{95}            & \underline{95.2}      \\\cline{3-25}
& \multicolumn{1}{|c|}{}                         &  FPFH+knn                           & \underline{89}            & 95.3          & \underline{94.1}          & 90.1          & \underline{93.5}          & 92.6          & 94.7          & \underline{95.9}          & 94.7          & 95.9          & 93.8          & 96.4          & 87.9          & 96.1          & 93.3          & \underline{93.9}          & 95            & 95.5          & 91.4          & 93.5          & 93.7          & 93.6   \\\cline{3-25}
& \multicolumn{1}{|c|}{\multirow{-4}{*}{ADD(S)}} &  DGCNN~\cite{wang2019dgcnn}         & 87.2          & 93.4          & 92            & 90.4          & 88.9          & 94.5          & 92            & 93.5          & 90.4          & 94.4          & 91.8          & 90.9          & 80.2          & 92.7          & 92.9          & 88.7          & 92.5          & 94            & 89.8          & 91.1          & 91.3          & 91.1   \\\cline{3-25}
& \multicolumn{1}{|c|}{\multirow{-4}{*}{AUC}}    &  SubdivNet~\cite{hu2021subdivnet}   &  89.1         & 91.6          & 90.2          & 89.7          & 90.4          & 91.6          & 92.6          & 94.4          & 91.5          & 94.2          & 91.4          & 88.8          & 80            & 93.3          & 91.8          & 89.2          & 91.9          & 91.1          & 90.2          & 89            & 91.6          & 90.6       \\\cline{2-25}\hline
\end{tabular}
\end{adjustbox}
\caption{YCB ADD-S AUC~\cite{posecnn} and ADD(S) AUC~\cite{hinterstoisser2012model} results with different descriptors: Non-symmetric objects are evaluated with ADD, and
symmetric objects
(annotated with~$^{*}$) are evaluated with ADD-S for ADD(S) AUC. The ADD-S AUC metrics is based on the curve with ADD-S for both non-symmetries and symmetries.
\label{tab:YCBVideoDescriptors}
}
\end{center}

\end{sidewaystable*}

%% file: figs/fig-resultsmore.tex
\begin{figure*}[h!]
\begin{center}
  \includegraphics[width=\textwidth]{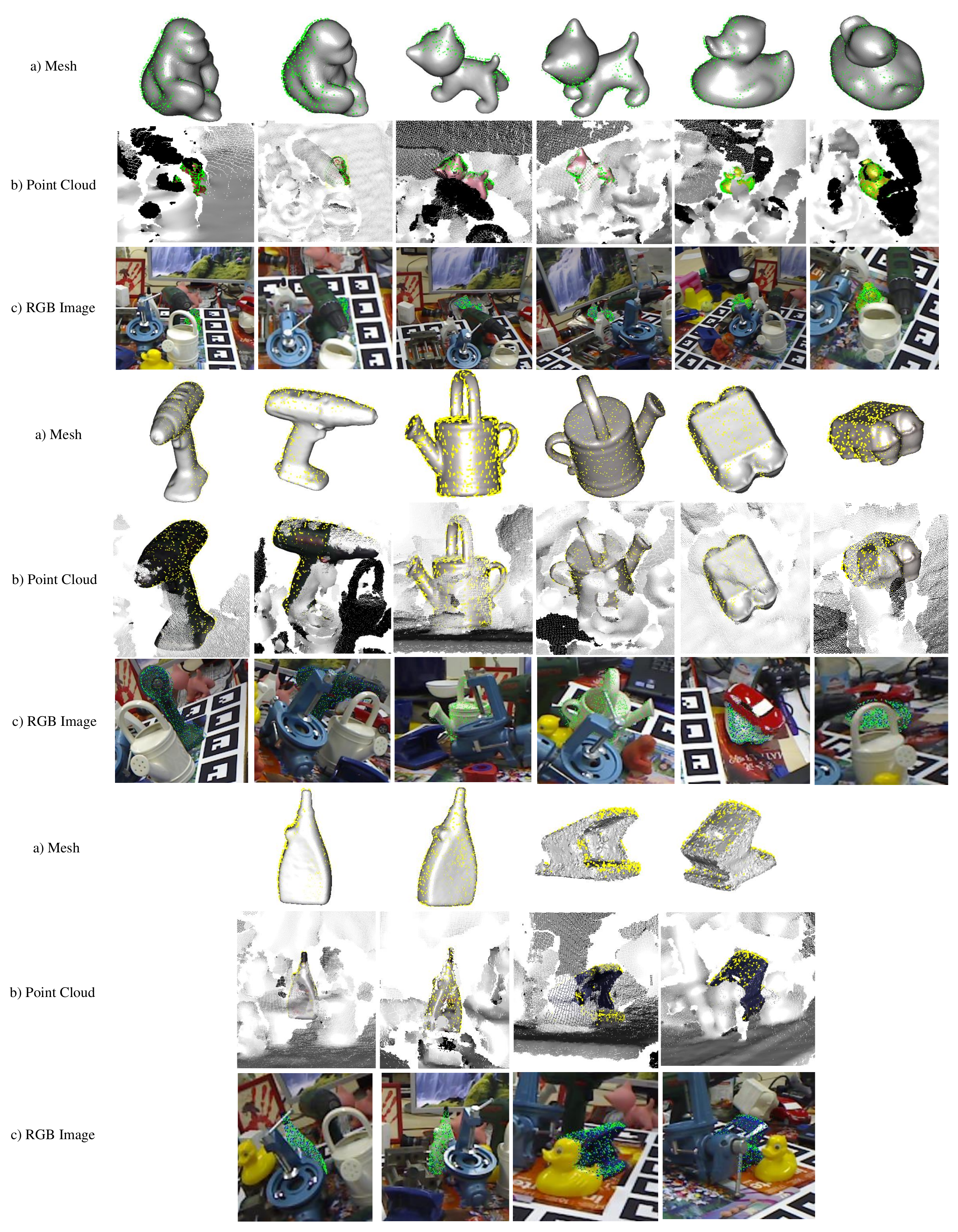}
  \vspace{-1.5\baselineskip}
  \caption{More Samples of RCVPose3D results on Occlusion LINEMOD: The meshes are applied with groundtruth (GT) pose, the green points are applied with estimated poses, whereas the blue dots are projected GT poses. The color in Point Cloud and RGB images are for illustration only, as RGB data is not used for training or inference} 
  \label{fig:resultsmore}
\end{center}
\vspace{0\baselineskip}
\end{figure*}